\documentclass[10pt,twocolumn,letterpaper]{article}

\usepackage[accsupp]{axessibility}  
\usepackage{cvpr}              
\definecolor{cvprblue}{rgb}{0.21,0.49,0.74}
\usepackage[pagebackref,breaklinks,colorlinks,allcolors=cvprblue]{hyperref}

\usepackage{multirow}
\usepackage{graphicx}
\usepackage{amsmath}
\usepackage{amssymb}
\usepackage{booktabs}
\usepackage{bm}

\usepackage{pifont}
\usepackage{xcolor}         
\usepackage{colortbl}
\usepackage{makecell}
\usepackage{amssymb}
\definecolor{demphcolor}{RGB}{144,144,144}
\definecolor{mygray}{gray}{0.95}
\usepackage{subcaption}
\usepackage{wrapfig} 
\def\paperID{} 
\def\confName{CVPR}
\def\confYear{2026}

\title{Demo2Tutorial: From Human Experience to Multimodal Software Tutorials}

\author{Zechen Bai,
Zhiheng Chen,
Yiqi Lin,
Kevin Qinghong Lin, \\
Difei Gao,
Xiangwu Guo,
Xin Wang,
Mike Zheng Shou\textsuperscript{\ding{41}} \\
Show Lab, National University of Singapore
}

\begin{document}
\maketitle
\def\thefootnote{\ding{41}}\footnotetext{Corresponding Author.}

\begin{abstract}
Human experience in digital environments offers a vast, underexplored resource of authentic, untrimmed interactions that contain rich procedural knowledge.
We introduce Demo2Tutorial, a framework that transforms this experience captured via screen recordings and interaction logs into structured, multimodal software tutorials for teaching both humans and agents.
Demo2Tutorial first collects human experience via a dedicated recorder, then parses raw experience using a multimodal Action Parser to reconstruct perception, action, and intent.
A Step Planner then abstracts these steps into hierarchical task graphs representing goals and steps.
Finally, a Tutorial Composer transforms the parsed experience into structured, reusable image-text instructions.
We evaluate the tutorial generation quality on a new benchmark derived from official software documentation.
We further demonstrate that this distilled representation benefits (i) human learning, by automatically generating multimodal tutorials, and (ii) agent learning, by improving downstream GUI-agent planning and generalization.
Experiments show Demo2Tutorial produces high-quality tutorials that surpass human-authored ones and significantly outperform baseline methods, while enabling both faster human task completion and improved GUI agent planning, demonstrating that structured tutorials distilled from human experience can serve as effective knowledge representations for advancing both human learning and agent capabilities.
Code and data will be available at \url{https://github.com/showlab/Demo2Tutorial}.
\end{abstract}
    
\section{Introduction}
\label{sec:intro}

\begin{figure}[t]
    \centering
    \includegraphics[width=\linewidth]{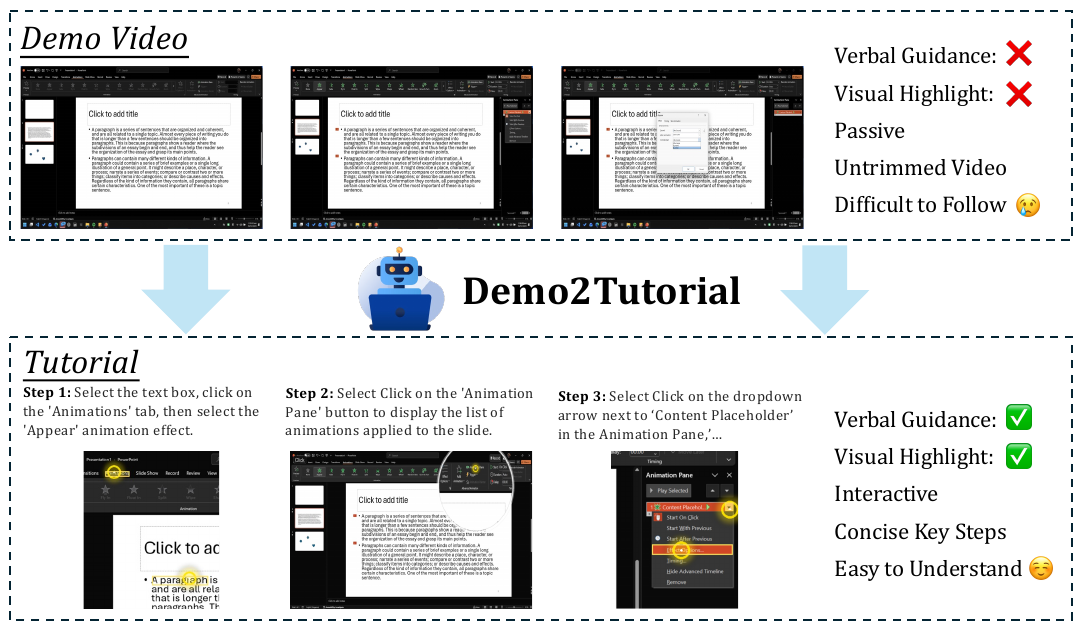}
    \caption{\textbf{Demonstration vs. Tutorial.} Raw demonstration videos are passive, untrimmed (redundant) recordings that lack verbal guidance and visual highlights, making them difficult to follow. In contrast, tutorials provide interactive, step-by-step instructions with clear verbal guidance and visual annotations, making software learning easy to understand.}
    \label{fig:teaser}
    \vspace{-5mm}
\end{figure}

Human experience represents a vast repository of procedural knowledge~\cite{schaal1996learning,chang2020procedure,grauman2024ego,grauman2022ego4d,early_experience}, which captures authentic, untrimmed interactions about what humans see, decide, and do.
Despite this progress in real-world instructional video understanding~\cite{li2022bridge,zhong2023learning,ashutosh2023video,mavroudi2023learning,nagasinghe2024not}, the extension to \emph{interactive digital environment} remains largely unexplored.
In this paper, we are interested in how to effectively extract this procedural knowledge from raw human interaction experience in digital environment. 
The goal is to distill this knowledge into a reusable representation that serves a dual purpose: enabling humans to efficiently learn new digital skills, such as mastering unfamiliar software, and empowering computer-use agents to more effectively accomplish everyday tasks in desktop environments.

Human experience in using computers mainly manifests in two forms:
raw screen recordings, often termed ``demonstrations,'' and tailored, step-by-step ``tutorials''.
While they may appear to cover similar content, a critical gap exists between them, as shown in Fig.~\ref{fig:teaser}.
From an audience perspective, the primary purpose of a demonstration is to showcase functionality, \ie, ``what it does'', treating the audience as passive observers.
In contrast, the main goal of a tutorial is to teach a process or skill, \ie, ``how to do it'', engaging audiences as active participants.
Regarding presentation format, demonstrations are often continuous, passive viewing experiences, frequently lacking verbal or visual guidance.
Tutorials, however, are explicitly structured for learning, complete with step-by-step instructions and highlights.
In concept, a tutorial takes a significant step beyond a mere demonstration by transforming raw action into digestible, actionable knowledge.
As a highly informative representation, image-text interleaved tutorials effectively condense human experience into reusable knowledge that can benefit both human learners and downstream GUI agents.

Despite the clear advantages of tutorials, their creation is often a labor-intensive process.
For instance, in a desktop environment, creating a demonstration can be as simple as recording the screen.
In contrast, curating a high-quality tutorial from this recording presents several significant challenges.
The first is long context compression: one must effectively select and summarize key steps from a long raw demonstration, which may be replete with verbose or irrelevant actions.
The second is the need for multimodal guidance: a qualified tutorial should present each step multimodally, combining clear verbal narration with visual highlights (e.g., zoom-in) to direct the user's attention.
These challenges typically require significant human effort, far exceeding that of recording a simple demonstration.

Towards automating this process, we introduce \textbf{Demo2Tutorial}, an agentic framework designed to automatically collect human computer-use experience and convert it into a structured, multimodal tutorial document.
This document comprises procedure narrations, concrete actions, and visual illustrations.
The Demo2Tutorial framework includes four core components:
(1) A Human Experience Recorder (HE-Recorder), a dedicated screen recorder that captures the screen and low-level human operations (\eg, clicks, keystrokes) simultaneously.
(2) An Action Parser that parses the semantics of these low-level actions.
(3) A Step Planner that abstracts the entire workflow, organizing the task and sub-goals into a hierarchical graph structure while filtering out irrelevant operations.
(4) A Tutorial Composer that outputs the final interleaved image-text content for each sub-goal in the task graph.
This involves both text explanation generation and intelligent image editing, such as zoom-ins and highlights, for visual clarity.
These components work in concert to deliver pleasant and meaningful tutorials.

In our evaluation, we first assess the quality of the automatically generated tutorials.
To this end, we curate a new benchmark, \textbf{TutorialBench}, which includes 110 tutorials covering 7 common software applications.
These tutorials are collected from the official websites of the software, serving as the human standard for quality.
We design dedicated evaluation metrics tailored for tutorial quality assessment across two domains and five dimensions. The \textit{Content Score} evaluates three key dimensions: Actionability, Completeness, and Conciseness.
The \textit{Visual Score} assesses two dimensions: Annotation quality and Image Relevance.
These metrics enable comprehensive evaluation of both the textual and visual aspects of generated tutorials.

To further study the utility of our distilled knowledge, we conduct two downstream evaluations on its benefits for both agent and human learning.
First, we evaluate the benefit for GUI agents by integrating generated tutorials into the Agent-S3~\cite{agents3} framework and testing on the OSWorld~\cite{xie2024osworld} benchmark,
Our results demonstrate substantial performance gains,
showing that multimodal tutorials provide effective external knowledge for planning and interface reasoning.
Second, we validate the benefit for human learning through a comparative user study where participants complete software tasks using either raw video demonstrations or generated tutorials.
We measure task completion time and direct preference between the two formats, with results showing that tutorials reduce learning time and are strongly preferred by users, confirming the effectiveness of our approach for both human and machine learners.

In summary, our contributions are threefold:
(1) We propose Demo2Tutorial, a novel agentic framework that automatically distills raw human computer-use experience into structured, multimodal tutorials.
(2) We introduce TutorialBench, a new benchmark for evaluating the quality of generated software tutorials.
(3) We demonstrate that our generated tutorials benefit both GUI agents (enhancing planning performance) and humans (efficient software learning).

\section{Related Works}

\noindent \textbf{Visual Design Automation}
Recent advances in multimodal foundation models~\cite{llava,showo,bai2024hallucination,bai2024one} have driven significant progress in automating various visual design tasks.
Several works tackle the challenge of converting academic papers into visual posters.
Paper2Poster~\cite{pang2025paper2poster} introduces a benchmark and multi-agent framework to distill papers into structured visual assets.
PosterGen~\cite{zhang2025postergen} and PosterForest~\cite{choi2025posterforest} further advance this by incorporating aesthetic principles and hierarchical representations.
Presentation automation has also seen notable progress with systems like PPTAgent~\cite{zheng2025pptagent}, which generates and evaluates presentations beyond simple text-to-slide conversion, and AutoPresent~\cite{ge2025autopresent}, which designs structured visuals from scratch. Other works such as D2S~\cite{sun2021d2s}, and SlideSpawn~\cite{kumar2024slidespawn} explore various approaches to automated slide deck creation.
While these works automate visual design from structured documents, our work addresses a distinct challenge: transforming raw, unstructured human demonstrations into pedagogical tutorials.
This requires capabilities beyond existing visual design systems, which primarily optimize for aesthetic quality.

\begin{figure*}[t!]
    \centering
    \includegraphics[width=0.95\linewidth]{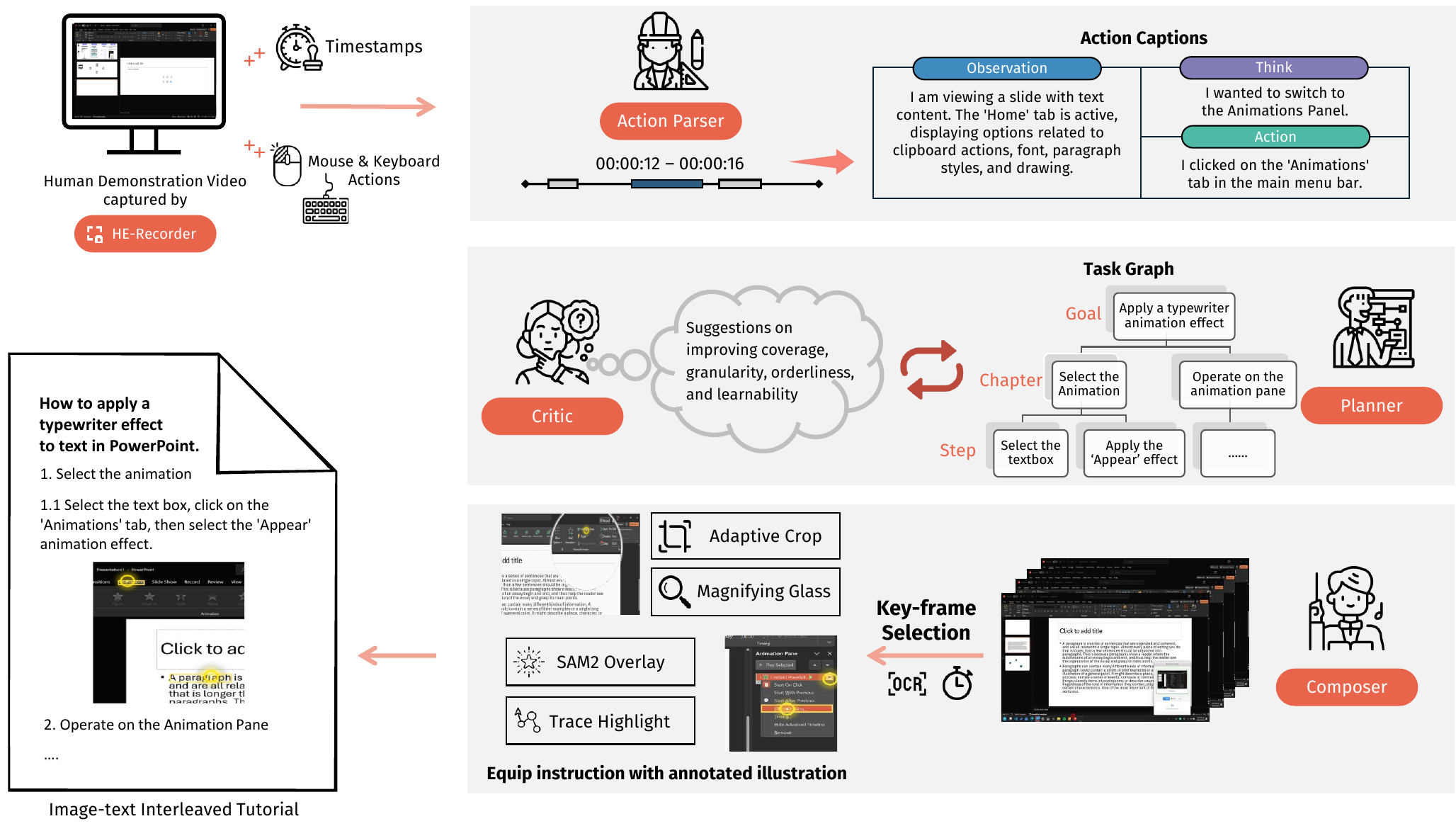}
    \vspace{-3mm}
    \caption{
    \textbf{Overview of Demo2Tutorial.} 
    Our framework comprises four key components: 
    (1) \textit{HE-Recorder} captures synchronized screen video and user actions in desktop environments. 
    (2) \textit{Action Parser} analyzes the recorded data using VLM-based semantic parsing to generate natural language descriptions of observations, actions, and user intents. 
    (3) \textit{Step Planner} organizes parsed actions into hierarchical task graphs through actor-critic iterative refinement, where the Planner generates structured tutorial drafts and the Critic provides feedback for quality improvement. 
    (4) \textit{Tutorial Composer} transforms the refined task graph into image-text interleaved tutorials with intelligent key-frame selection, adaptive visual highlights and annotations for enhanced accessibility.
    }
    \vspace{-2mm}
    \label{fig:demo2tutorial}
\end{figure*}

\noindent \textbf{Computer Use Agents}
Recent advances in computer-use agents formulate GUI automation as a visual-language-action problem.
Representative works like AssistGUI~\cite{gao2024assistgui}, ShowUI~\cite{lin2025showui}, UI-TARS~\cite{qin2025ui,wang2025ui}, and OpenCUA~\cite{wang2025opencua} collectively advance computer-use automation by unifying vision, language, and action in end-to-end models that directly predict executable GUI action from human instruction and observed screenshot input.
Concurrently, several studies~\cite{agents2,agents3,song2025coact,yang2025gta1} explore agentic frameworks that integrate multiple specialized experts, such as grounding, planning, and coding, within iterative workflows following the ReAct~\cite{yao2022react} paradigms, enabling agents to reason, act, and self-evaluate in a unified loop.
A complementary line of research explores learning from human demonstrations.
OpenCUA~\cite{wang2025opencua} collects large-scale human-screen interaction trajectories to support model pre-training by imitating human demonstrations. VideoAgentTrek~\cite{lu2025videoagenttrek} extends this paradigm by mining unlabeled screen-recording videos to bootstrap model learning without explicit demonstrations.
VideoWebArena~\cite{jang2024videowebarena} further evaluates long-context multimodal agents on web tasks with video understanding, highlighting the importance of representing interaction experience in a form that agents can effectively consume.

While prior work focuses on directly training agents from raw demonstrations, our work takes a complementary approach: we transform human demonstrations into structured, reusable tutorials that serve as external knowledge for both human learners and GUI agents.
This distillation process enables instruction following rather than implicit behavior imitation, providing interpretable guidance that can enhance agent planning and facilitate human learning.
\section{Demo2Tutorial}

Our Demo2Tutorial framework transforms raw human demonstrations into structured, pedagogical tutorials (Fig.~\ref{fig:demo2tutorial}).
It first captures synchronized visual and action streams via a dedicated recorder, then parses these low-level actions into semantic descriptions, organizes them into hierarchical tutorial structures through iterative refinement by planner and critic, and finally composes the output into multimodal tutorials with intelligent visual annotations.

\subsection{Human Experience Recorder}
\label{sec:he_recorder}

The Human Experience Recorder (HE-Recorder) serves as the foundation of our framework, capturing the complete spectrum of human computer-use experience in desktop environments.
Unlike existing screen recorders that only capture visual streams, HE-Recorder records both high-fidelity screen video and low-level user actions (clicks, keystrokes, movements) with precise temporal synchronization.
This dual-stream recording is crucial for downstream semantic understanding and tutorial generation, distinguishing our approach from methods relying solely on visual observations.
The recorder comprises three key components:

\noindent \textbf{1) Video Recording.}
We employ FFmpeg with the \texttt{ddagrab} filter to capture full-screen, full-resolution video at 30 FPS.
This is natively implemented in the recorder without requiring external software, \eg, OBS.

\noindent \textbf{2) Action Recording.}
Building upon KeyCastOW~\footnote{\url{https://github.com/brookhong/KeyCastOW}}, a C++-based keystroke visualizer, we implement real-time action logging that captures all mouse operations (clicks, movements, drags) and keyboard inputs with high-precision timestamps.
Unlike the original visualization purpose, our implementation outputs action sequences to a structured log file for subsequent semantic parsing.

\noindent \textbf{3) Time Synchronization.}
To address inevitable timing delays between video and action streams across different machines, we implement an interactive calibration mechanism.
At recording start, an on-screen timer prompts users to press a hotkey, establishing a common temporal reference point.
This synchronization step is essential for aligning dense action sequences with corresponding video frames, particularly critical when capturing expert users performing rapid, consecutive operations.

\subsection{Action Parser}
\label{sec:action_parser}

The Action Parser transforms raw action logs and visual observations into semantically rich, natural language descriptions in a bottom-up manner.

\noindent \textbf{1) Data Calibration.}
Raw action logs undergo three preprocessing steps to ensure quality and consistency.
\textit{i)} we calibrate time offsets using the hotkey timestamp from HE-Recorder as a pivot to synchronize actions with video frames.
\textit{ii)} consecutive keystrokes within 1-second windows are merged into single ``typing'' actions to reduce granularity while preserving semantic meaning.
\textit{iii)} redundant modifier keys (Shift, Ctrl, Alt) are consolidated into ``Shortcut Key'' actions when applicable.

\noindent \textbf{2) VLM-based Semantic Parsing.}
We employ GPT-4o to parse each action into natural language descriptions.
For each action, we extract before-and-after screenshots from the video and apply action-grounded visual prompting: drawing red bounding boxes at mouse coordinates to highlight interaction regions~\cite{clip_red_circle,yang2023som}.
To improve parsing accuracy and mitigate hallucinations, we design a Chain-of-Thought prompting strategy to instruct the VLM to output five sequential fields: (1) observation before, (2) observation after, (3) differences between states, (4) factual action description, and (5) inferred user intent.
This structured reasoning helps the VLM ground actions in visual changes and distinguishes low-level operations and high-level intentions, which is crucial for subsequent task-level abstraction.

\subsection{Step Planner}

Given the parsed action sequence, the Step Planner organizes low-level actions into a hierarchical task graph structure.
Unlike top-down approaches that decompose tasks into subtasks, we employ a bottom-up strategy: first grouping atomic actions into meaningful steps, then clustering related steps into chapters, and finally synthesizing an overarching tutorial goal.
This design naturally filters irrelevant operations while preserving the authentic workflow from human demonstrations.

\noindent \textbf{Hierarchical Abstraction.}
The planner transforms a flat action sequence into a three-level hierarchy.
At the \textit{step level}, consecutive actions addressing a common sub-goal (\eg, ``adjust font size to 24pt'') are grouped and abstracted into a single instructional step with an action verb (click/type/select/drag).
At the \textit{chapter level}, semantically related steps are clustered into logical phases (\eg, ``Basic Configuration,'' ``Content Editing''), each representing a distinct stage of the workflow.
Finally, the \textit{tutorial goal} is synthesized from all chapters to capture the overall task objective.
For long sequences, we apply chunking at natural breakpoints (software switches, temporal gaps) to maintain tractable LLM context, then merge chunk outputs while resolving temporal and hierarchical dependencies.

\noindent \textbf{Actor-Critic Iterative Refinement.}
To ensure pedagogical quality, we implement an actor-critic loop where the Planner acts as the \textit{actor} and a separate Critic agent evaluates the generated tutorial draft.
In each iteration, the Planner generates a structured tutorial JSON with the hierarchical organization and step-level instructions.
The Critic then evaluates this draft across multiple dimensions (e.g., coverage, granularity, orderliness, learnability) and provides actionable feedback if quality thresholds are not met.
The Planner incorporates this feedback in the next iteration, refining instruction clarity, adjusting granularity, or reorganizing chapter boundaries.
This loop continues until the Critic approves the draft or the maximum iteration limit is reached, ensuring both structural coherence and instructional quality.

\subsection{Tutorial Composer}

The Tutorial Composer converts the structured tutorial draft generated by the iterative refinement into an image-text interleaved format suitable for both human presentation and multimodal model training.
It orchestrates a suite of perception and editing tools to automatically reconstruct visual narratives for every instructional step, ensuring both pedagogical clarity and visual fidelity.

\noindent \textbf{Key-frame Selection.}
A key technical design of the Composer is its intelligent frame selection mechanism.
For each instructional step, rather than uniformly sampling frames or selecting by fixed timestamps, the Composer samples multiple candidate frames within a temporal window around the action and evaluates them using a multi-dimensional scoring function.
This function combines four weighted criteria: (1) \textit{text relevance}: semantic alignment between frame content and instruction text via OCR-based matching; (2) \textit{image sharpness}: Laplacian variance to avoid motion blur; (3) \textit{motion stability}: temporal consistency to prevent mid-transition captures; and (4) \textit{temporal proximity}: Gaussian-weighted distance from the action timestamp.
The highest-scoring frame is selected as the key visual anchor, ensuring optimal visual quality and semantic grounding even for complex multi-step operations or drag-and-drop actions.
This score-based selection mechanism, powered by perception tools like \textit{RapidOCR} and visual analysis, significantly outperforms naive sampling strategies.

\noindent \textbf{Adaptive Visual Highlight.}
Once key-frames are selected, the Composer actively applies editing operations to enhance visual clarity.
Leveraging perception tools such as \textit{SAM2} for UI component segmentation and \textit{RapidOCR} for text region detection, the Composer dynamically overlays action-specific visual annotations: click markers for point interactions, drag trajectories for movement operations, hotkey badges for keyboard shortcuts, and magnifier effects for fine-grained details.
Beyond annotation, the system employs intelligent editing functions—adaptive cropping to focus on action-dense regions, highlighting to emphasize relevant UI elements, and contrast adjustment to improve visibility.
These editing operations are contextually applied based on action type and interface layout, ensuring that each tutorial step provides clear, unambiguous visual guidance.

\noindent The integrated outputs are compiled into a structured interleaved dataset presented in both \textit{markdown} and \textit{JSON} formats, supporting intuitive human comprehension as well as downstream multimodal fine-tuning.

\section{TutorialBench}
\label{sec:benchmark}

To quantitatively evaluate Demo2Tutorial's performance against a human standard, we curated TutorialBench, a dedicated benchmark comprising 110 samples.
These samples span seven widely-used daily software applications.
Each sample in the benchmark is a triplet: $\langle$\textit{Task Goal}, \textit{Raw Demonstration}, \textit{Official Tutorial}$\rangle$.
This structure allows the agent to use the raw demonstration and task goal (optional) as input, while the official, human-authored tutorial serves as the reference for evaluation.

\noindent \textbf{Tutorial Collection.}
We selected seven popular software applications: Microsoft Office (Word, PowerPoint, Excel) and Adobe Creative Suite (Acrobat, Premiere Pro, Photoshop, After Effects).
These applications were chosen for their widespread daily usage and the availability of dedicated high-quality tutorials on the official websites.
We sourced our ground-truth tutorials directly from these official websites, following a strict set of collection criteria.
We required each tutorial: 1) represent a non-trivial task where a tutorial provides tangible value;
2) can be presented in the interleaved image-text format, which is the target output for our agent; and 3) focus on a single, self-contained task that can be clearly described in a one-sentence goal statement.

\noindent \textbf{Demonstration Collection.}
To collect the raw demonstration inputs, we recruited human experts familiar with these software applications.
The experts used our HE-Recorder to record their screen and actions.
We implemented a specific protocol to ensure the demonstrations captured natural, goal-directed behavior rather than a mechanical copy.

First, the expert was shown both the task goal and the official tutorial to fully understand the required procedure.
Then, to prevent a direct 1:1 copy, the expert was explicitly prohibited from viewing the tutorial again while recording the demonstration.
This forces the expert to execute the task from their own procedural understanding, resulting in a more realistic and untrimmed workflow, complete with the natural pauses or minor detours that characterize the authentic human experience.

\begin{figure}[t]
    \centering
    \includegraphics[width=\linewidth]{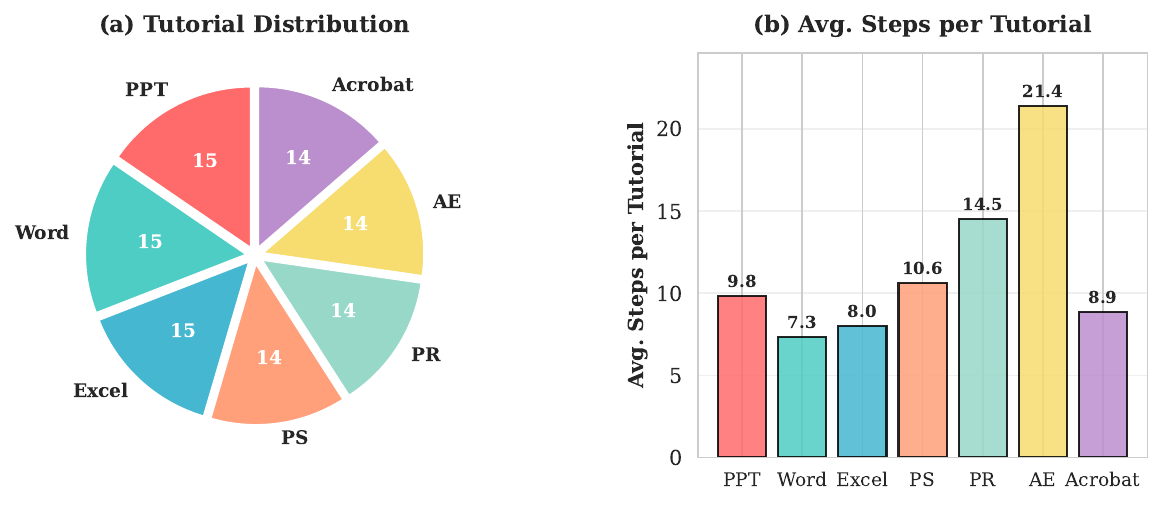}
    \vspace{-2mm}
    \caption{\textbf{TutorialBench statistics across 7 software applications.} (a) Distribution of 110 tutorials across applications. (b) Average steps of human authored tutorials per software, with After Effects tutorials being most complex (21.4 avg steps) and Word tutorials being most concise (7.35 avg steps).}
    \label{fig:bench_stats}
    \vspace{-4mm}
\end{figure}

\noindent \textbf{Data Statistics.}
Overall, TutorialBench comprises 110 samples.
The dataset is balanced across the 7 applications: PowerPoint (17), Word (17), Excel (16), Photoshop (15), Premiere Pro (15), After Effects (15), and Acrobat (15).
Further statistics, including the distribution of tutorials across applications and the number of steps per tutorial, are detailed in Fig.~\ref{fig:bench_stats}.

\section{Experiment}

We conduct comprehensive experiments to assess Demo2Tutorial from three complementary perspectives.
First, we evaluate the intrinsic quality of generated tutorials against human-authored references in TutorialBench using VLM-as-judge metrics across content and visual dimensions, comparing our framework with end-to-end and multi-agent baselines.
Second, we investigate whether generated tutorials serve as effective external knowledge for GUI agents by integrating them into the Agent-S3 planning framework and measuring task success rates on OSWorld benchmark.
Third, we validate the pedagogical utility for human learners through a comparative user study, measuring task completion time and format preference between raw demonstrations and generated tutorials.

\begin{table*}[t]
\centering
\caption{\textbf{Tutorial Generation Quality Evaluation on TutorialBench.} We compare our framework (Demo2Tutorial) against the ground truth (GT) and various baseline approaches across Content Score (Actionability, Completeness, Conciseness) and Visual Score (Annotation, Image Relevance) dimensions. Best results in each column are highlighted in \textbf{bold}.}
\vspace{-3mm}
\label{tab:main_results}
\resizebox{0.85\linewidth}{!}{
\begin{tabular}{l|ccc|c|cc|c|c}
\toprule
& \multicolumn{4}{c|}{\textbf{Content Score}} & \multicolumn{3}{c|}{\textbf{Visual Score}} & \\
\cmidrule(lr){2-5} \cmidrule(lr){6-8}
\textbf{Framework} & Action. & Complete. & Concise. & Avg. & Annot. & Img Rel. & Avg. & \textbf{Overall} \\
\midrule
GT (Human) & 81.0 & 90.6 & 83.1 & 84.9 & 54.4 & 86.6 & 70.5 & 79.1 \\
\midrule
\multicolumn{9}{l}{\textit{End-to-End Methods}} \\
\midrule
Text-based Generation & 75.4 & 62.0 & 40.1 & 59.2 & -- & -- & -- & -- \\
Vision-based Generation & 78.2 & \textbf{95.1} & 65.1 & 79.4 & 9.2 & 74.0 & 41.6 & 64.3 \\
\midrule
\multicolumn{9}{l}{\textit{Multi-Agent Methods}} \\
\midrule
Vanilla Multi-Agent & 71.1 & 88.9 & 59.0 & 73.0 & 51.3 & 81.5 & 66.4 & 70.3 \\
\rowcolor{blue!15}
\textbf{Demo2Tutorial} & \textbf{90.5} & 92.3 & \textbf{70.8} & \textbf{84.5} & \textbf{83.3} & \textbf{94.0} & \textbf{88.7} & \textbf{86.2} \\
\bottomrule
\end{tabular}
}
\vspace{-3mm}
\end{table*}

\begin{table}[ht]
\centering
\caption{OSWorld Subset Experiment Results}
\vspace{-3mm}
\label{tab:os_world}
\renewcommand{\arraystretch}{1}
\setlength{\tabcolsep}{8pt}
\resizebox{0.82\linewidth}{!}{
\begin{tabular}{lcc}
    \toprule
    \textbf{Model Setup} & \textbf{Chrome (17)} & \textbf{VLC (14)} \\
    \midrule
    \rowcolor{gray!15} o4-mini (baseline)     & 47.1 & 53.4 \\
    \textit{+Tutorial} & \textbf{58.8} & \textbf{56.1} \\
    $\Delta$ & \textit{(+11.7$\uparrow$)} & \textit{(+2.7$\uparrow$)} \\
    \hline
    \rowcolor{gray!15} GPT-5 (baseline) & 52.9 & 59.6 \\
    \textit{+Tutorial} & \textbf{70.6} & \textbf{70.7} \\
    $\Delta$ & \textit{(+17.6$\uparrow$)} & \textit{(+11.1$\uparrow$)} \\
    \bottomrule
\end{tabular}
}
\vspace{-5mm}
\end{table}

\subsection{Evaluating Tutorial Generation Quality}

We evaluate the quality of generated tutorials using our proposed two-domain, five-dimension metrics (Sec.~\ref{sec:intro}): \textit{Content Score} (Actionability, Completeness, Conciseness) and \textit{Visual Score} (Annotation quality, Image Relevance).
Evaluation is performed on TutorialBench using VLM-as-judge (GPT-4o) to score each dimension on a 0 to 1 scale (multiplied by 100 for better readability in the table).
We further validate the reliability of VLM-as-Judge via a human consistency study, where the averaged human scores correlate well with VLM-as-Judge (\(\rho=0.755\)).
Additional details, including the correlation table, runtime/cost statistics, and ablations, are provided in the Supplementary.

\noindent\textbf{Comparison Methods.}

\textit{GT (Human)} serves as our reference, comprising official tutorials sourced from software vendor websites as described in Sec.~\ref{sec:benchmark}.
Notably, we observe that these human-authored tutorials often provide screenshots for only a subset of steps, lacking comprehensive visual annotations throughout, a reflection of the labor-intensive nature of manual tutorial creation.

For \textit{end-to-end methods}, we evaluate two variants: 
(1) \textit{Text-based Generation} operates as a ``blind'' baseline, using only the Action Parser output (action descriptions) to conduct pure text-based generation without visual input or output.
(2) \textit{Vision-based Generation} frames the problem as video understanding, uniformly sampling a fixed number of frames from the raw demonstration video and prompting the model to directly output the tutorial.

\textit{Vanilla Multi-Agent} makes full use of HE-Recorder by providing both action logs and corresponding key-frames as input, then prompting GPT to generate tutorials.
Compared to Vision-based methods, this approach ensures frames are action-aligned key-frames rather than uniform samples.
However, unlike Demo2Tutorial, it lacks advanced agentic components, specifically the Actor-Critic iterative refinement and the Composer's adaptive visual highlighting.

\noindent\textbf{Results and Analysis.}
Tab.~\ref{tab:main_results} shows Demo2Tutorial achieves an overall score of 86.2, surpassing human-authored tutorials (79.1) and all baselines.

\textit{End-to-end methods} struggle with tutorial generation. Text-based Generation scores only 59.2 overall, demonstrating that visual grounding is essential for creating effective tutorials.
Text alone cannot convey interface interactions.
Vision-based Generation achieves high Completeness (95.1) by including numerous sampled frames, yet suffers catastrophically in visual annotation (9.2), resulting in an overall score of 64.3.
This reveals that uniform sampling produces visually cluttered, pedagogically weak tutorials without meaningful visual guidance.

\textit{Vanilla Multi-Agent} (70.3) benefits from action-aligned key-frames but still lags behind Demo2Tutorial by 15.9.
This gap highlights the critical contributions of our actor-critic refinement loop, which iteratively improves instruction clarity and structural coherence, and the Composer's intelligent visual annotation, which adaptively highlights relevant UI elements.

Notably, Demo2Tutorial excels in Visual Score (88.7 vs. 70.5 for GT), demonstrating that our automated key-frame selection and adaptive visual annotation strategies surpass manual tutorial creation in visual presentation quality.
While GT tutorials maintain slightly higher Completeness through manual curation that may include supplementary information, Demo2Tutorial achieves superior Actionability, indicating clearer, more executable instructions that better serve both human and GUI agent learners.

\begin{figure}
    \centering
    \includegraphics[width=0.85\linewidth]{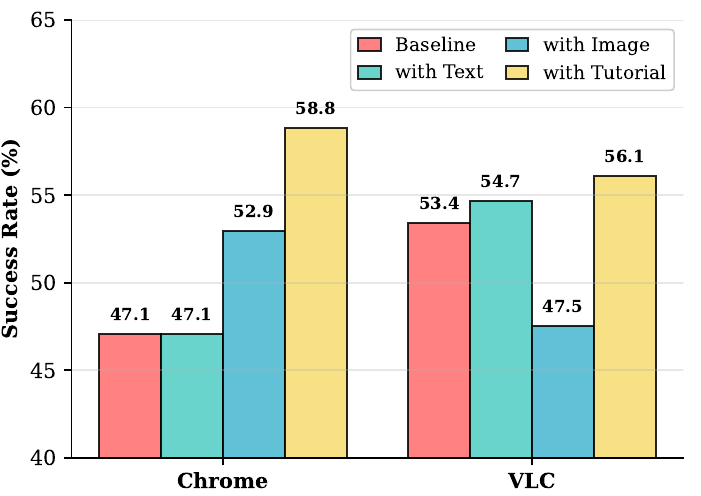}
    \vspace{-3mm}
    \caption{Ablation study of Tutorial in OSWorld Subset.}
    \label{fig:osworld_ab}
\end{figure}

\subsection{Tutorials Teach GUI Agent}

In this section, we explore whether multimodal tutorials can serve as an effective source of external knowledge to enhance the planning capability of GUI agents in real-world computer-use environments. 
To this end, we evaluate two representative planning models under the Agent-S3~\cite{agents3} framework on the \textbf{OSWorld}~\cite{xie2024osworld} benchmark.

\noindent \textbf{Experimental Setup.}
Two strong planning models, \textit{GPT-o4-mini} and \textit{GPT-5}, were evaluated on two representative domains: \textit{Chrome} and \textit{VLC}. 
Each agent was tasked with executing standardized desktop manipulation procedures drawn from OSWorld’s benchmark suite. 
Performance is reported as the mean success rate across 17 Chrome and 14 VLC tasks.
We compare four levels of contextual supervision:
(a) \textit{Baseline} (prompt-only),
(b) \textit{with Text} (textual content from tutorials),
(c) \textit{with Image} (visual content from tutorials), and
(d) \textit{with Tutorial} (full content).

\noindent \textbf{Main Results.}
\cref{tab:os_world} reports the quantitative results across different model and domains. 
For \textit{GPT-o4-mini}, incorporating tutorial signals improves success rates from \textbf{47.1\%} to \textbf{58.8\%} in Chrome and from \textbf{53.4\%} to \textbf{56.1\%} in VLC. 
The stronger \textit{GPT-5} model exhibits even stronger gains, rising from \textbf{52.9\%} to \textbf{70.6\%} in Chrome and from \textbf{59.6\%} to \textbf{70.7\%} in VLC. 
These consistent improvements suggest that tutorials as explicit knowledge facilitate better planning stability during GUI task execution.

\noindent \textbf{Ablation Analysis.}
\cref{fig:osworld_ab} visualizes the ablation over the four context baselines for \textit{GPT-o4-mini}. 
Adding textual guidance (\textit{+Text}) produces only marginal improvement, implying that linguistic information lacks sufficient grounding for visual actions reasoning. 
Visual cues (\textit{+Image}) contribute variably, helpful for Chrome but less stable in VLC, possibly due to domain-specific visual variance. 
In contrast, multimodal tutorials (\textit{+Tutorial}) consistently yield the highest success rates, reinforcing that tightly coupled language–vision alignment provides the strongest knowledge enhancement for action sequencing and interface reasoning.

\subsection{Tutorials Teach Human}

To validate the pedagogical utility of generated tutorials for human learners, we conduct a comparative user study where participants learn and complete a software task using either raw demonstration videos or generated tutorials.

\begin{figure}[t]
    \centering
    \includegraphics[width=\linewidth]{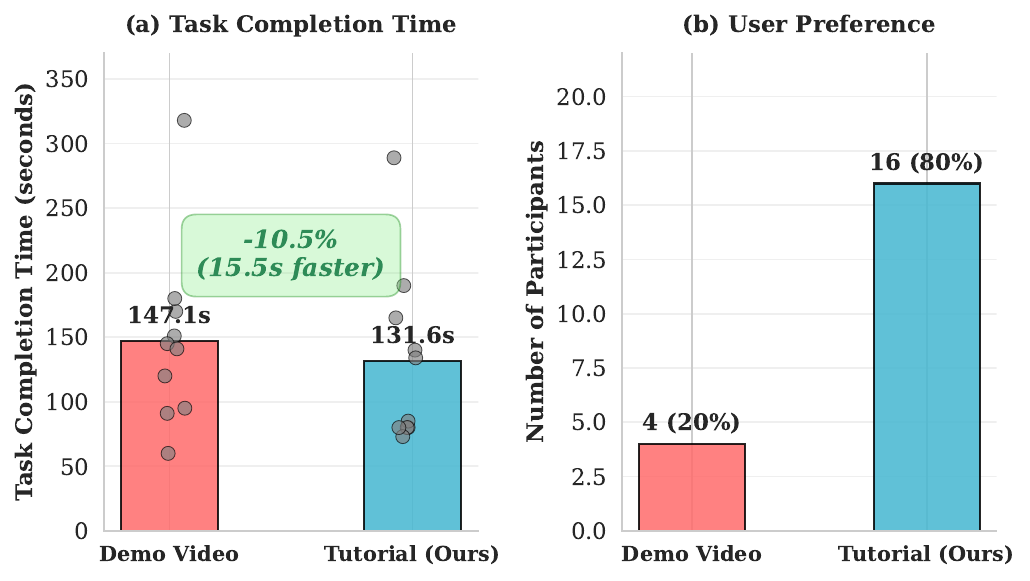}
    \vspace{-3mm}
    \caption{\textbf{Human user study results.} (a) Task completion time: tutorials enable 10.5\% faster completion. (b) User preference: 80\% of participants prefer tutorials over raw demonstration videos for learning software operations.}
    \vspace{-3mm}
    \label{fig:user_study_results}
\end{figure}

\begin{figure*}[t!]
    \centering
    \includegraphics[width=0.97\textwidth]{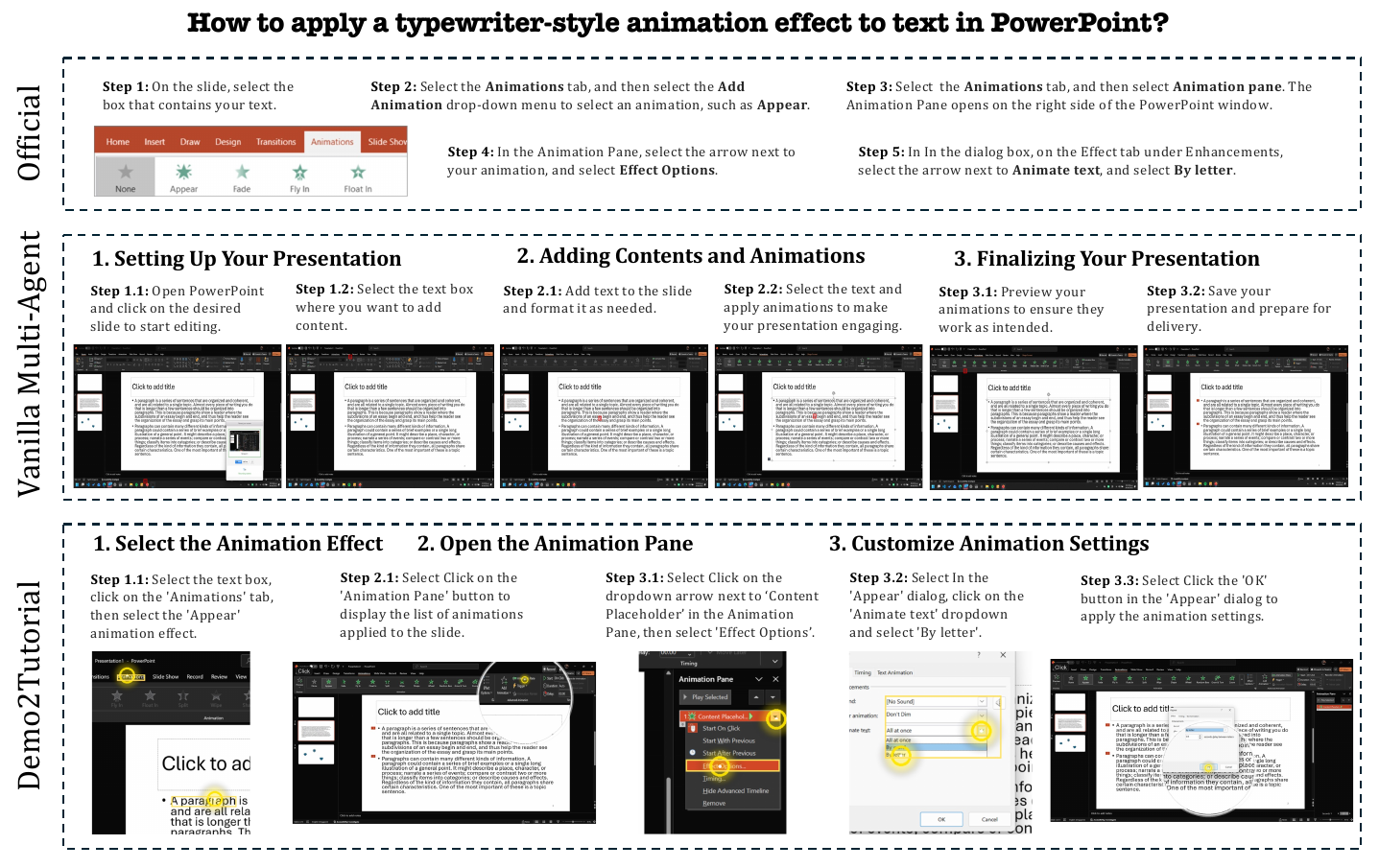}
    \vspace{-3mm}
    \caption{\textbf{Qualitative comparison between official tutorial (top), agent baseline (middle) and Demo2Tutorial output (bottom).} Official tutorials typically use minimal screenshots with dense text descriptions, requiring users to locate UI elements themselves.
    Agent baseline fails to produce semantically aligned image-text pairs and lacks visual highlights.
    Our generated tutorials provide step-by-step visual grounding with adaptive annotations that explicitly show where to interact, significantly improving clarity and learnability.}
    \label{fig:qualitative}
    \vspace{-5mm}
\end{figure*}

\noindent\textbf{Experimental Setup.}
We recruit 20 participants and randomly assign them into two groups (10 per condition).
The task involves implementing a seldom-used animation effect in Microsoft PowerPoint, specifically, creating a custom motion path animation with timing adjustments.
This task is intentionally chosen to be non-trivial, requiring multiple steps that are not immediately obvious to novice users.
Participants in the \textit{Demo Video} condition watch the raw screen recording demonstration, while those in the \textit{Tutorial} condition study the image-text interleaved tutorial generated by Demo2Tutorial.
After the learning phase, participants attempt to complete the task independently while being timed.
All participants successfully completed the task, confirming that both learning materials are effective.
After task completion, participants answer a preference question: ``Which format do you find more suitable for learning this skill?''

\noindent\textbf{Results.}
As shown in Fig.~\ref{fig:user_study_results}(a), participants using tutorials complete the task in 131.6 seconds on average, \textbf{10.5\% faster} than those using demo videos (147.1s).
This time reduction demonstrates that tutorials enable more efficient learning, \ie, participants can quickly locate and understand relevant steps through structured text instructions and annotated visuals, whereas video viewers must scrub through the timeline to find key moments.

More importantly, user preference strongly favors tutorials: \textbf{16 out of 20 participants (80\%)} prefer the tutorial format over raw videos for learning software operations (Fig.~\ref{fig:user_study_results}(b)).
This strong preference, combined with faster task completion, validates that automatically generated tutorials provide superior pedagogical value compared to raw demonstration videos.

\subsection{Qualitative Examples}

We present a representative comparison in Fig.~\ref{fig:qualitative} showing three tutorial versions for creating a custom animation in PowerPoint: the official tutorial from Microsoft, Vanilla Multi-Agent, and Demo2Tutorial.

The official tutorial provides only a single screenshot with dense text descriptions, forcing users to mentally map instructions to the interface.
Vanilla Multi-Agent, despite having access to action logs and key-frames, produces vague instructions (\eg, ``apply animations'') that lack actionable guidance.
Moreover, its screenshots suffer from image-text misalignment, frames do not correspond to the actions and completely lack visual annotations, resulting in plain screenshots that provide minimal visual guidance.

In contrast, Demo2Tutorial generates comprehensive image-text interleaved tutorials where each operation is paired with a semantically aligned screenshot selected through our scoring mechanism.
Critically, our adaptive visual annotations, click markers, region highlights, and UI element emphasis, provide explicit visual grounding that directs users' attention to relevant interface components.
This comparison validates the importance of our key technical designs: key-frame selection ensures image-text alignment, while adaptive visual highlighting transforms plain screenshots into pedagogically effective instructional materials.

\section{Conclusion}
\label{sec:conclusion}

We present Demo2Tutorial, an agentic framework that automatically transforms raw human computer-use demonstrations into structured, multimodal tutorials.
Through synchronized experience recording, VLM-based action parsing, actor-critic iterative refinement, and adaptive visual composition, our framework generates high-quality tutorials that surpass human-authored ones (86.2 vs. 79.1 overall score on TutorialBench) while significantly outperforming end-to-end and multi-agent baselines.
We demonstrate dual utility: tutorials improve GUI agent planning on OSWorld (GPT-5: 52.9\%→70.6\% success rate in Chrome), and enable 10.5\% faster human task completion with 80\% user preference over raw videos.
By bridging raw demonstrations and pedagogical tutorials, Demo2Tutorial establishes a scalable path for converting human expertise into reusable knowledge that benefits both human learners and intelligent agents.
Future work includes extending to mobile/web platforms, improving computational efficiency, and enabling personalized tutorial generation.

\section*{Acknowledgements}
This research is supported by the National Research Foundation, Singapore under its AI Singapore Programme (AISG Award No: AISG3-RP-2022-030).

{
    \small
    \bibliographystyle{ieeenat_fullname}
    \bibliography{main}

@inproceedings{lin2025showui,
  title={Showui: One vision-language-action model for gui visual agent},
  author={Lin, Kevin Qinghong and Li, Linjie and Gao, Difei and Yang, Zhengyuan and Wu, Shiwei and Bai, Zechen and Lei, Stan Weixian and Wang, Lijuan and Shou, Mike Zheng},
  booktitle={Proceedings of the Computer Vision and Pattern Recognition Conference},
  pages={19498--19508},
  year={2025}
}

@article{qin2025ui,
  title={Ui-tars: Pioneering automated gui interaction with native agents},
  author={Qin, Yujia and Ye, Yining and Fang, Junjie and Wang, Haoming and Liang, Shihao and Tian, Shizuo and Zhang, Junda and Li, Jiahao and Li, Yunxin and Huang, Shijue and others},
  journal={arXiv preprint arXiv:2501.12326},
  year={2025}
}

@article{wang2025ui,
  title={Ui-tars-2 technical report: Advancing gui agent with multi-turn reinforcement learning},
  author={Wang, Haoming and Zou, Haoyang and Song, Huatong and Feng, Jiazhan and Fang, Junjie and Lu, Junting and Liu, Longxiang and Luo, Qinyu and Liang, Shihao and Huang, Shijue and others},
  journal={arXiv preprint arXiv:2509.02544},
  year={2025}
}

@article{wang2025opencua,
  title={Opencua: Open foundations for computer-use agents},
  author={Wang, Xinyuan and Wang, Bowen and Lu, Dunjie and Yang, Junlin and Xie, Tianbao and Wang, Junli and Deng, Jiaqi and Guo, Xiaole and Xu, Yiheng and Wu, Chen Henry and others},
  journal={arXiv preprint arXiv:2508.09123},
  year={2025}
}

@article{agents2,
  title={Agent s2: A compositional generalist-specialist framework for computer use agents},
  author={Agashe, Saaket and Wong, Kyle and Tu, Vincent and Yang, Jiachen and Li, Ang and Wang, Xin Eric},
  journal={arXiv preprint arXiv:2504.00906},
  year={2025}
}

@article{agents3,
  title={The Unreasonable Effectiveness of Scaling Agents for Computer Use},
  author={Gonzalez-Pumariega, Gonzalo and Tu, Vincent and Lee, Chih-Lun and Yang, Jiachen and Li, Ang and Wang, Xin Eric},
  journal={arXiv preprint arXiv:2510.02250},
  year={2025}
}

@article{song2025coact,
  title={Coact-1: Computer-using agents with coding as actions},
  author={Song, Linxin and Dai, Yutong and Prabhu, Viraj and Zhang, Jieyu and Shi, Taiwei and Li, Li and Li, Junnan and Savarese, Silvio and Chen, Zeyuan and Zhao, Jieyu and others},
  journal={arXiv preprint arXiv:2508.03923},
  year={2025}
}

@article{yang2025gta1,
  title={Gta1: Gui test-time scaling agent},
  author={Yang, Yan and Li, Dongxu and Dai, Yutong and Yang, Yuhao and Luo, Ziyang and Zhao, Zirui and Hu, Zhiyuan and Huang, Junzhe and Saha, Amrita and Chen, Zeyuan and others},
  journal={arXiv preprint arXiv:2507.05791},
  year={2025}
}

@article{lu2025videoagenttrek,
  title={VideoAgentTrek: Computer Use Pretraining from Unlabeled Videos},
  author={Lu, Dunjie and Xu, Yiheng and Wang, Junli and Wu, Haoyuan and Wang, Xinyuan and Wang, Zekun and Yang, Junlin and Su, Hongjin and Chen, Jixuan and Chen, Junda and others},
  journal={arXiv preprint arXiv:2510.19488},
  year={2025}
}

@inproceedings{yao2022react,
  title={React: Synergizing reasoning and acting in language models},
  author={Yao, Shunyu and Zhao, Jeffrey and Yu, Dian and Du, Nan and Shafran, Izhak and Narasimhan, Karthik R and Cao, Yuan},
  booktitle={The eleventh international conference on learning representations},
  year={2022}
}

@article{xie2024osworld,
  title={Osworld: Benchmarking multimodal agents for open-ended tasks in real computer environments},
  author={Xie, Tianbao and Zhang, Danyang and Chen, Jixuan and Li, Xiaochuan and Zhao, Siheng and Cao, Ruisheng and Hua, Toh J and Cheng, Zhoujun and Shin, Dongchan and Lei, Fangyu and others},
  journal={Advances in Neural Information Processing Systems},
  volume={37},
  pages={52040--52094},
  year={2024}
}

@article{pang2025paper2poster,
  title={Paper2Poster: Towards Multimodal Poster Automation from Scientific Papers},
  author={Pang, Wei and Lin, Kevin Qinghong and Jian, Xiangru and He, Xi and Torr, Philip},
  journal={arXiv preprint arXiv:2505.21497},
  year={2025}
}

@article{zhang2025postergen,
  title={Postergen: Aesthetic-aware paper-to-poster generation via multi-agent llms},
  author={Zhang, Zhilin and Zhang, Xiang and Wei, Jiaqi and Xu, Yiwei and You, Chenyu},
  journal={arXiv preprint arXiv:2508.17188},
  year={2025}
}

@article{choi2025posterforest,
  title={PosterForest: Hierarchical Multi-Agent Collaboration for Scientific Poster Generation},
  author={Choi, Jiho and Park, Seojeong and Song, Seongjong and Shim, Hyunjung},
  journal={arXiv preprint arXiv:2508.21720},
  year={2025}
}

@inproceedings{zheng2025pptagent,
  title={Pptagent: Generating and evaluating presentations beyond text-to-slides},
  author={Zheng, Hao and Guan, Xinyan and Kong, Hao and Zhang, Wenkai and Zheng, Jia and Zhou, Weixiang and Lin, Hongyu and Lu, Yaojie and Han, Xianpei and Sun, Le},
  booktitle={Proceedings of the 2025 Conference on Empirical Methods in Natural Language Processing},
  pages={14413--14429},
  year={2025}
}

@inproceedings{ge2025autopresent,
  title={Autopresent: Designing structured visuals from scratch},
  author={Ge, Jiaxin and Wang, Zora Zhiruo and Zhou, Xuhui and Peng, Yi-Hao and Subramanian, Sanjay and Tan, Qinyue and Sap, Maarten and Suhr, Alane and Fried, Daniel and Neubig, Graham and others},
  booktitle={Proceedings of the Computer Vision and Pattern Recognition Conference},
  pages={2902--2911},
  year={2025}
}

@article{sun2021d2s,
  title={D2S: Document-to-slide generation via query-based text summarization},
  author={Sun, Edward and Hou, Yufang and Wang, Dakuo and Zhang, Yunfeng and Wang, Nancy XR},
  journal={arXiv preprint arXiv:2105.03664},
  year={2021}
}

@article{kumar2024slidespawn,
  title={Slidespawn: An automatic slides generation system for research publications},
  author={Kumar, Keshav and Chowdary, Ravindranath},
  journal={arXiv preprint arXiv:2411.17719},
  year={2024}
}

@inproceedings{clip_red_circle,
  title={What does clip know about a red circle? visual prompt engineering for vlms},
  author={Shtedritski, Aleksandar and Rupprecht, Christian and Vedaldi, Andrea},
  booktitle={Proceedings of the IEEE/CVF International Conference on Computer Vision},
  pages={11987--11997},
  year={2023}
}

@article{yang2023som,
  title={Set-of-mark prompting unleashes extraordinary visual grounding in gpt-4v},
  author={Yang, Jianwei and Zhang, Hao and Li, Feng and Zou, Xueyan and Li, Chunyuan and Gao, Jianfeng},
  journal={arXiv preprint arXiv:2310.11441},
  year={2023}
}

@article{schaal1996learning,
  title={Learning from demonstration},
  author={Schaal, Stefan},
  journal={Advances in neural information processing systems},
  volume={9},
  year={1996}
}

@article{early_experience,
  title={Agent Learning via Early Experience},
  author={Zhang, Kai and Chen, Xiangchao and Liu, Bo and Xue, Tianci and Liao, Zeyi and Liu, Zhihan and Wang, Xiyao and Ning, Yuting and Chen, Zhaorun and Fu, Xiaohan and others},
  journal={arXiv preprint arXiv:2510.08558},
  year={2025}
}

@inproceedings{chang2020procedure,
  title={Procedure planning in instructional videos},
  author={Chang, Chien-Yi and Huang, De-An and Xu, Danfei and Adeli, Ehsan and Fei-Fei, Li and Niebles, Juan Carlos},
  booktitle={European Conference on Computer Vision},
  pages={334--350},
  year={2020},
  organization={Springer}
}

@inproceedings{grauman2024ego,
  title={Ego-exo4d: Understanding skilled human activity from first-and third-person perspectives},
  author={Grauman, Kristen and Westbury, Andrew and Torresani, Lorenzo and Kitani, Kris and Malik, Jitendra and Afouras, Triantafyllos and Ashutosh, Kumar and Baiyya, Vijay and Bansal, Siddhant and Boote, Bikram and others},
  booktitle={Proceedings of the IEEE/CVF Conference on Computer Vision and Pattern Recognition},
  pages={19383--19400},
  year={2024}
}

@inproceedings{grauman2022ego4d,
  title={Ego4d: Around the world in 3,000 hours of egocentric video},
  author={Grauman, Kristen and Westbury, Andrew and Byrne, Eugene and Chavis, Zachary and Furnari, Antonino and Girdhar, Rohit and Hamburger, Jackson and Jiang, Hao and Liu, Miao and Liu, Xingyu and others},
  booktitle={Proceedings of the IEEE/CVF conference on computer vision and pattern recognition},
  pages={18995--19012},
  year={2022}
}

@inproceedings{li2022bridge,
  title={Bridge-prompt: Towards ordinal action understanding in instructional videos},
  author={Li, Muheng and Chen, Lei and Duan, Yueqi and Hu, Zhilan and Feng, Jianjiang and Zhou, Jie and Lu, Jiwen},
  booktitle={Proceedings of the IEEE/CVF conference on computer vision and pattern recognition},
  pages={19880--19889},
  year={2022}
}

@inproceedings{zhong2023learning,
  title={Learning procedure-aware video representation from instructional videos and their narrations},
  author={Zhong, Yiwu and Yu, Licheng and Bai, Yang and Li, Shangwen and Yan, Xueting and Li, Yin},
  booktitle={Proceedings of the IEEE/CVF Conference on Computer Vision and Pattern Recognition},
  pages={14825--14835},
  year={2023}
}

@article{ashutosh2023video,
  title={Video-mined task graphs for keystep recognition in instructional videos},
  author={Ashutosh, Kumar and Ramakrishnan, Santhosh Kumar and Afouras, Triantafyllos and Grauman, Kristen},
  journal={Advances in Neural Information Processing Systems},
  volume={36},
  pages={67833--67846},
  year={2023}
}

@inproceedings{mavroudi2023learning,
  title={Learning to ground instructional articles in videos through narrations},
  author={Mavroudi, Effrosyni and Afouras, Triantafyllos and Torresani, Lorenzo},
  booktitle={Proceedings of the IEEE/CVF International Conference on Computer Vision},
  pages={15201--15213},
  year={2023}
}

@inproceedings{nagasinghe2024not,
  title={Why not use your textbook? knowledge-enhanced procedure planning of instructional videos},
  author={Nagasinghe, Kumaranage Ravindu Yasas and Zhou, Honglu and Gunawardhana, Malitha and Min, Martin Renqiang and Harari, Daniel and Khan, Muhammad Haris},
  booktitle={Proceedings of the IEEE/CVF Conference on Computer Vision and Pattern Recognition},
  pages={18816--18826},
  year={2024}
}

@article{jang2024videowebarena,
  title={Videowebarena: Evaluating long context multimodal agents with video understanding web tasks},
  author={Jang, Lawrence and Li, Yinheng and Zhao, Dan and Ding, Charles and Lin, Justin and Liang, Paul Pu and Bonatti, Rogerio and Koishida, Kazuhito},
  journal={arXiv preprint arXiv:2410.19100},
  year={2024}
}

@article{bai2024hallucination,
  title={Hallucination of multimodal large language models: A survey},
  author={Bai, Zechen and Wang, Pichao and Xiao, Tianjun and He, Tong and Han, Zongbo and Zhang, Zheng and Shou, Mike Zheng},
  journal={arXiv preprint arXiv:2404.18930},
  year={2024}
}

@article{bai2024one,
  title={One token to seg them all: Language instructed reasoning segmentation in videos},
  author={Bai, Zechen and He, Tong and Mei, Haiyang and Wang, Pichao and Gao, Ziteng and Chen, Joya and Liu, Lei and Zhang, Zheng and Shou, Mike Z},
  journal={Advances in Neural Information Processing Systems},
  volume={37},
  pages={6833--6859},
  year={2024}
}

@inproceedings{gao2024assistgui,
  title={Assistgui: Task-oriented pc graphical user interface automation},
  author={Gao, Difei and Ji, Lei and Bai, Zechen and Ouyang, Mingyu and Li, Peiran and Mao, Dongxing and Wu, Qinchen and Zhang, Weichen and Wang, Peiyi and Guo, Xiangwu and others},
  booktitle={Proceedings of the IEEE/CVF Conference on Computer Vision and Pattern Recognition},
  pages={13289--13298},
  year={2024}
}

@article{llava,
  title={Visual instruction tuning},
  author={Liu, Haotian and Li, Chunyuan and Wu, Qingyang and Lee, Yong Jae},
  journal={Advances in neural information processing systems},
  volume={36},
  pages={34892--34916},
  year={2023}
}

@article{showo,
  title={Show-o: One single transformer to unify multimodal understanding and generation},
  author={Xie, Jinheng and Mao, Weijia and Bai, Zechen and Zhang, David Junhao and Wang, Weihao and Lin, Kevin Qinghong and Gu, Yuchao and Chen, Zhijie and Yang, Zhenheng and Shou, Mike Zheng},
  journal={arXiv preprint arXiv:2408.12528},
  year={2024}
}
}
\clearpage
\appendix

\section{More Analysis}

\subsection{Compression Analysis}

Essentially, the Demo2Tutorial pipeline is a knowledge compression process that transforms lengthy demonstrations into concise tutorials.
The compression happens in three key stages:
\begin{itemize}
    \item \textbf{Stage 1 (Action Parser):} HE-Recorder captures synchronized raw video frames and action logs. The Action Parser then performs action-based filtering, reducing thousands of raw frames to a compact sequence of trace steps by identifying and extracting only action-relevant frames.
    \item \textbf{Stage 2 (Step Planner):} The Planner performs semantic compression by abstracting the trace actions into hierarchical task graphs representing high-level goals and sub-goals, further reducing the number of steps while refining the semantic granularity.
    \item \textbf{Stage 3 (Tutorial Composer):} The Composer applies intelligent key-frame selection through multi-dimensional scoring (text relevance, image sharpness, motion stability, temporal proximity), producing the final compressed tutorial with optimal visual-textual alignment.
\end{itemize}

\begin{figure}[t]
    \centering
    \includegraphics[width=\linewidth]{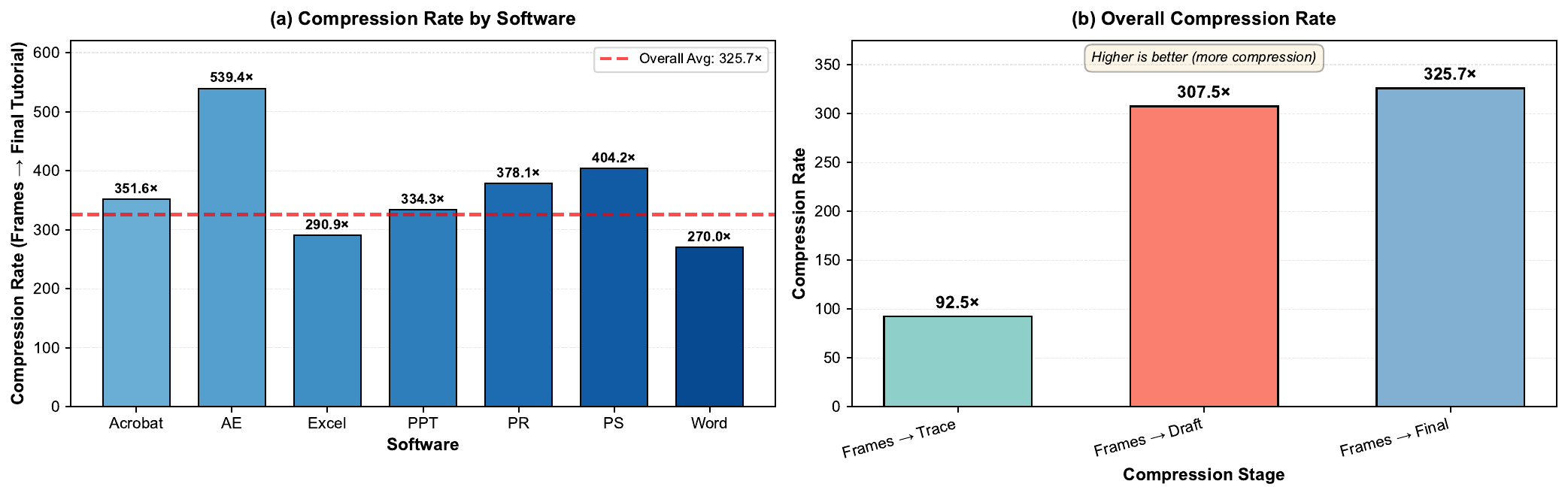}
    \caption{\textbf{Compression analysis across the Demo2Tutorial pipeline.} (a) Final compression rate (frames to final tutorial) for each software application, showing variation across different software complexity levels. After Effects achieves the highest compression (539.43×) while Word has the lowest (270.01×), with an overall average of 325.71×. (b) Stage-wise compression rates demonstrate that the majority of compression occurs in Stage 1 (Action Parser: 92.46×), followed by Stage 2 (Planner: 3.33×) and Stage 3 (Composer: 1.06×).}
    \label{fig:compression_analysis}
\end{figure}

Figure~\ref{fig:compression_analysis} quantifies the compression achieved across 110 demonstrations from TutorialBench.
On average, a demonstration contains 1,208 video frames (at 30 FPS, approximately 40 seconds), which is compressed to 13.07 trace steps by the Action Parser (92.46× compression), then to 3.93 draft steps by the Planner (additional 3.33× compression), and finally to 3.71 tutorial steps by the Composer (additional 1.06× compression), achieving an overall 325.71× compression ratio.
The compression rate varies across software applications: video editing software like After Effects (539.43×) and Premiere Pro (378.10×) achieve higher compression rates due to longer demonstration videos with repetitive operations, while productivity software like Word (270.01×) and Excel (290.93×) have lower compression rates as their demonstrations are typically shorter and more concise.
This multi-stage compression not only reduces storage and computational costs but also distills raw experience into pedagogically effective knowledge representations suitable for both human learning and agent training.

\subsection{Results with Different Backbone}

\begin{table*}[t]
\centering
\caption{Tutorial Generation Quality Evaluation with Different Backbone.}
\vspace{-3mm}
\label{tab:supp_qwen}
\resizebox{0.85\linewidth}{!}{
\begin{tabular}{l|ccc|c|cc|c|c}
\toprule
& \multicolumn{4}{c|}{\textbf{Content Score}} & \multicolumn{3}{c|}{\textbf{Visual Score}} & \\
\cmidrule(lr){2-5} \cmidrule(lr){6-8}
\textbf{Framework} & Action. & Complete. & Concise. & Avg. & Annot. & Img Rel. & Avg. & \textbf{Overall} \\
\midrule
GT (Human) & 81.0 & 90.6 & 83.1 & 84.9 & 54.4 & 86.6 & 70.5 & 79.1 \\
\midrule
Vanilla Multi-Agent (GPT-4o) & 71.1 & 88.9 & 59.0 & 73.0 & 51.3 & 81.5 & 66.4 & 70.3 \\
Demo2Tutorial (GPT-4o) & 90.5 & 92.3 & 70.8 & 84.5 & 83.3 & 94.0 & 88.7 & 86.2 \\
Demo2Tutorial (Qwen) & 87.2 & 82.0 & 50.0 & 73.1 & 85.4 & 92.7 & 89.1 & 79.5 \\
\bottomrule
\end{tabular}
}
\end{table*}

In the main paper, we use GPT-4o as the backbone model due to its strong multimodal capabilities and cost-effectiveness.
In Tab.~\ref{tab:supp_qwen}, we further evaluate our framework using Qwen-VL-32B-Instruct as an alternative backbone.
Results show that Qwen achieves lower overall performance than GPT-4o (79.5 vs. 86.2), with the most significant gap in the Conciseness dimension (50.0 vs. 70.8).
Despite this performance gap, Qwen's results remain competitive with human-authored tutorials (79.5 vs. 79.1), demonstrating that our agentic framework design can effectively enhance the capabilities of weaker MLLMs and produce high-quality tutorials even with less capable backbone models.

\subsection{Component Ablation on TutorialBench}
\label{sec:supp_ablation}

\begin{table*}[ht]
    \centering
    \scriptsize
    \setlength{\tabcolsep}{3pt}
    \caption{\textbf{Contribution of each component.} Ablations on a subset of TutorialBench using VLM-as-Judge scores (\(\times 100\)). The sharp degradation in Annotation Quality when removing Visual Highlight validates the necessity of adaptive visual guidance.}
    \label{tab:supp_component_contribution}
    \vspace{-3mm}
    \resizebox{0.85\linewidth}{!}{
    \begin{tabular}{@{}l|ccc|c|cc|c|c@{}}
        \toprule
        \textbf{Setting} & \textbf{Act.} & \textbf{Comp.} & \textbf{Conc.} & \textbf{Avg. Content} & \textbf{Annot.} & \textbf{ImgRel} & \textbf{Avg.Visual} & \textbf{Overall} \\
        \midrule
        GT (Human) & 77.6 & 90.5 & 84.8 & 84.3 & 45.2 & 86.7 & 66.0 & 77.0 \\
        Demo2Tutorial & 91.4 & 93.3 & 84.8 & 89.8 & 85.2 & 96.2 & 90.7 & 90.2 \\
        \midrule
        \textit{w/o} Task Hierarchy & 91.0 & 86.7 & 64.8 & 80.8 & 83.8 & 96.2 & 90.0 & 84.5 \\
        \textit{w/o} Visual Highlight & 85.2 & 79.5 & 73.3 & 79.4 & 3.8 & 89.0 & 46.4 & 66.2 \\
        \textit{w/o} Iterative Refinement & 91.0 & 91.9 & 75.2 & 86.0 & 85.2 & 95.2 & 90.2 & 87.7 \\
        \textit{w/o} Key-Frame Selection & 89.0 & 91.0 & 76.2 & 85.4 & 86.2 & 95.2 & 90.7 & 87.5 \\
        \bottomrule
    \end{tabular}
    }
    
    \vspace{-4mm}
\end{table*}

We perform component-wise ablations to quantify the contribution of each key design in Demo2Tutorial on a subset of TutorialBench.
The subset consists of 21 samples, 3 from each software respectively.
Specifically, we construct four variants by removing one component at a time under the same evaluation protocol:
\textit{(i)} \textit{w/o Task Hierarchy} removes hierarchical abstraction, producing flat step lists without chapter/goal structuring;
\textit{(ii)} \textit{w/o Visual Highlight} disables adaptive visual annotation and highlighting on key-frames;
\textit{(iii)} \textit{w/o Iterative Refinement} removes the actor-critic refinement loop in the Step Planner; and
\textit{(iv)} \textit{w/o Key-Frame Selection} replaces score-based key-frame selection with a simpler alternative without multi-factor scoring.
We then evaluate each variant using VLM-as-Judge on the same five dimensions.
Results are reported in Tab.~\ref{tab:supp_component_contribution}.

We observe that removing Visual Highlight causes a drastic collapse in Annotation Quality (85.2$\rightarrow$3.8) and Overall score (90.2$\rightarrow$66.2), showing that adaptive visual highlighting is essential for producing learnable tutorials.
Removing Task Hierarchy mainly degrades Conciseness (84.8$\rightarrow$64.8) and Overall (90.2$\rightarrow$84.5), indicating that hierarchical organization is crucial for avoiding verbose or poorly structured instructions.
Removing Iterative Refinement or Key-Frame Selection yields smaller but consistent drops in Overall score (90.2$\rightarrow$87.7/87.5), suggesting both refinement and frame selection improve instruction quality and image-text alignment.

\section{Experiment Details}

\subsection{VLM-as-Judge}

Evaluating tutorial quality presents unique challenges that traditional NLP metrics (e.g., BLEU, ROUGE) cannot adequately address.
First, tutorials are inherently multimodal.
Effective evaluation must jointly assess both textual instructions and visual components (screenshots, annotations).
Second, tutorial quality depends on pedagogical effectiveness rather than mere surface-level similarity to reference text: a tutorial with different wording but clearer instructions may be superior.
Third, human evaluation, while reliable, is prohibitively expensive and non-scalable for iterative development and large-scale benchmarking.
Recent advances in Vision-Language Models (VLMs) demonstrate strong capabilities in understanding multimodal content and making nuanced judgments, making them suitable candidates for automated tutorial evaluation.

\begin{figure*}[t!]
    \centering
    \includegraphics[width=0.97\textwidth]{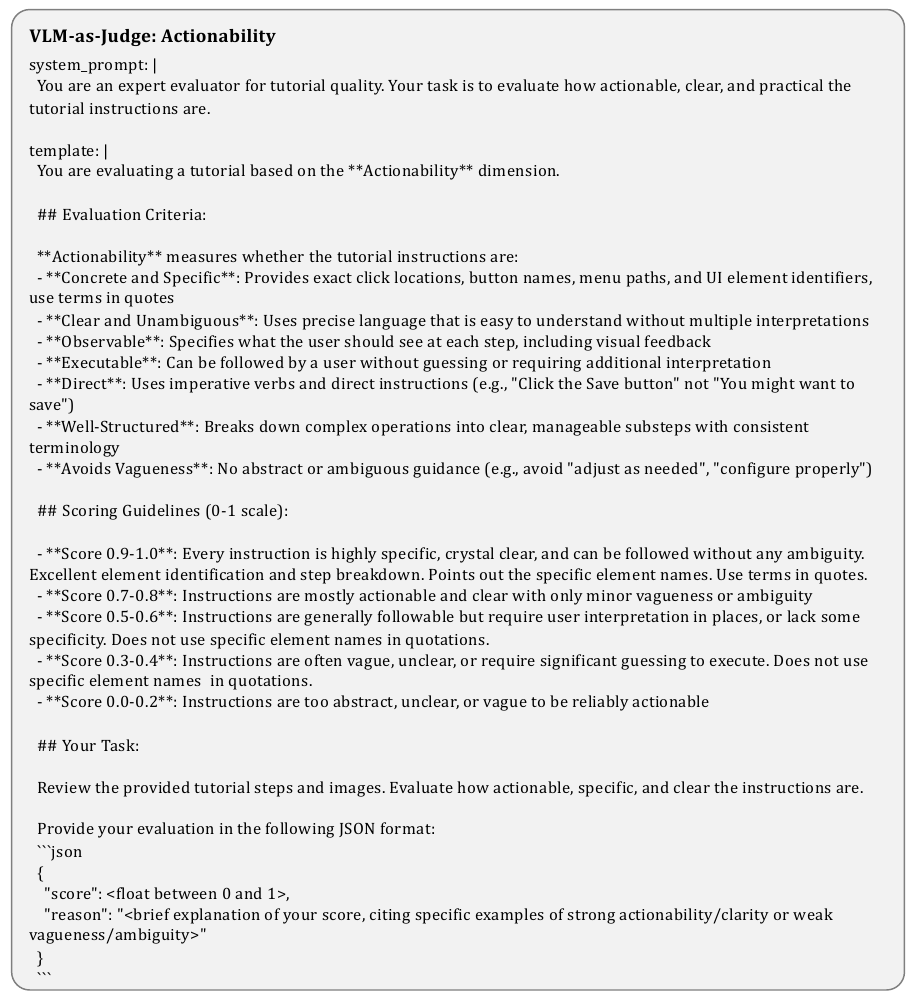}
    \caption{VLM-as-Judge prompt for Actionability.}
    \label{fig:vlm_judge_prompt_actionability}
\end{figure*}

\begin{figure*}[t!]
    \centering
    \includegraphics[width=0.97\textwidth]{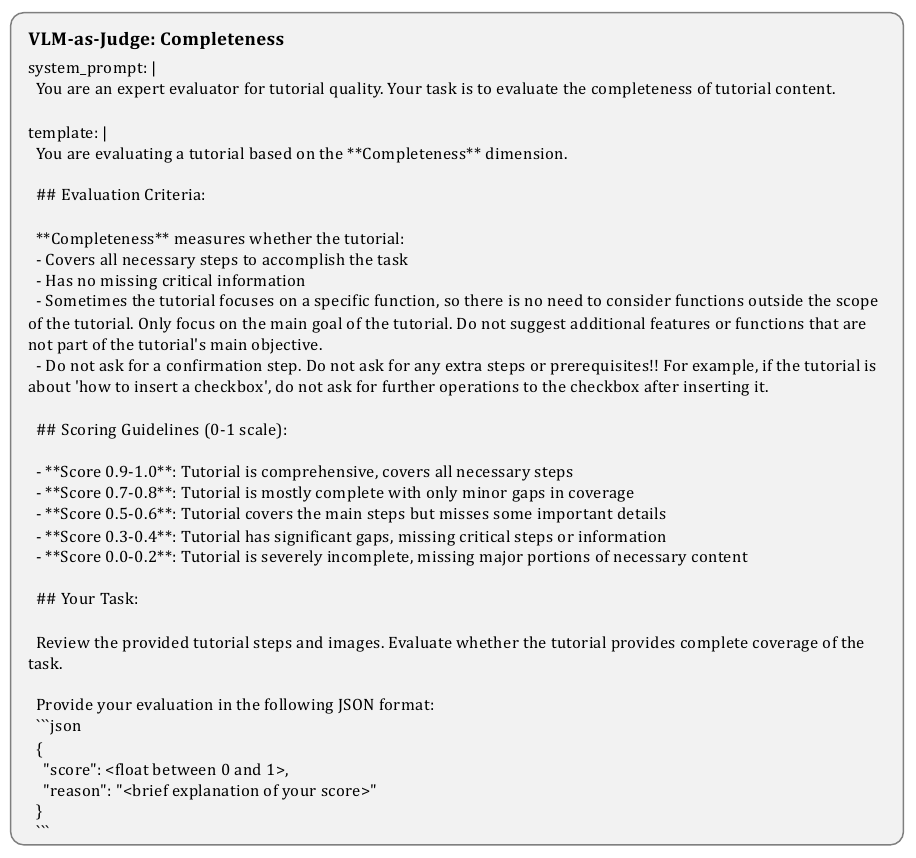}
    \caption{VLM-as-Judge prompt for Completeness.}
    \label{fig:vlm_judge_prompt_completeness}
\end{figure*}

\begin{figure*}[t!]
    \centering
    \includegraphics[width=0.97\textwidth]{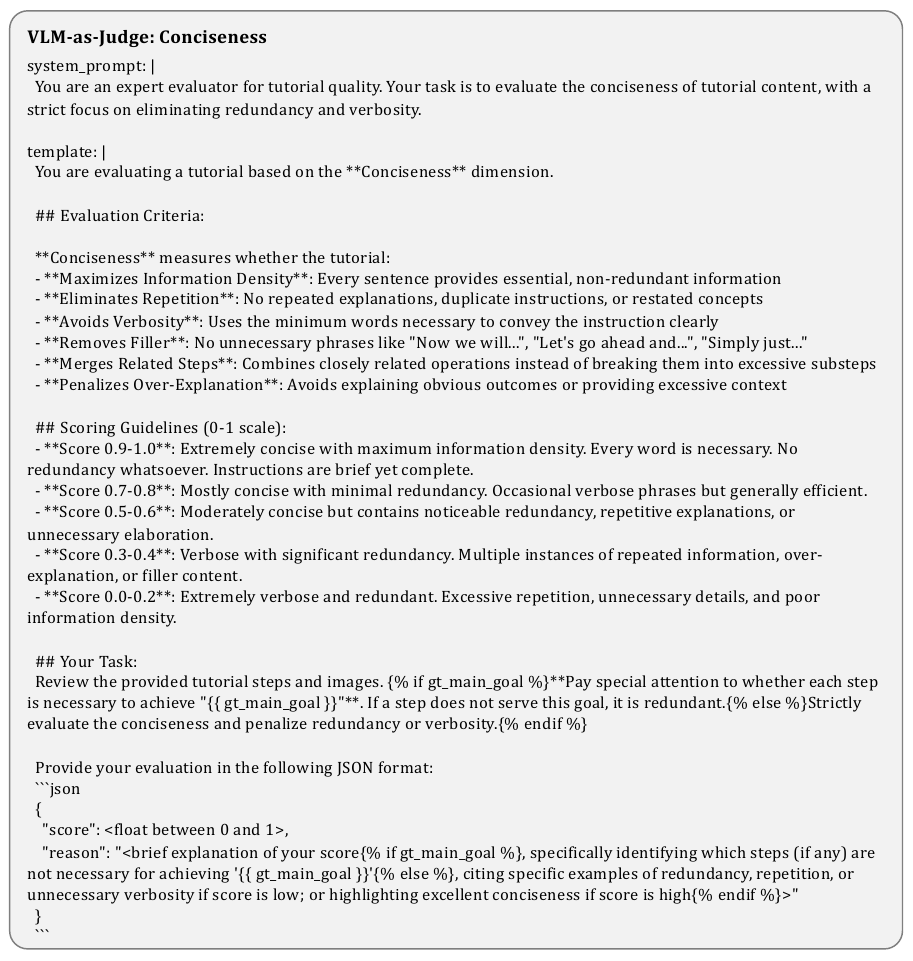}
    \caption{VLM-as-Judge prompt for Conciseness.}
    \label{fig:vlm_judge_prompt_conciseness}
\end{figure*}

Our VLM-as-Judge protocol is grounded in two core principles that capture the dual nature of software tutorials:

\textit{Content Score} evaluates the instructional quality of textual descriptions across three dimensions:
\begin{itemize}
    \item \textbf{Actionability}: Can users successfully execute the described operations based solely on the provided instructions? This measures clarity and specificity. See \cref{fig:vlm_judge_prompt_actionability} for the prompt.
    \item \textbf{Completeness}: Are all necessary steps included without missing critical operations? This measures information coverage. See \cref{fig:vlm_judge_prompt_completeness} for the prompt.
    \item \textbf{Conciseness}: Are instructions clear and direct without unnecessary verbosity? This measures pedagogical efficiency. See \cref{fig:vlm_judge_prompt_conciseness} for the prompt.
\end{itemize}

\begin{figure*}[t!]
    \centering
    \includegraphics[width=0.97\textwidth]{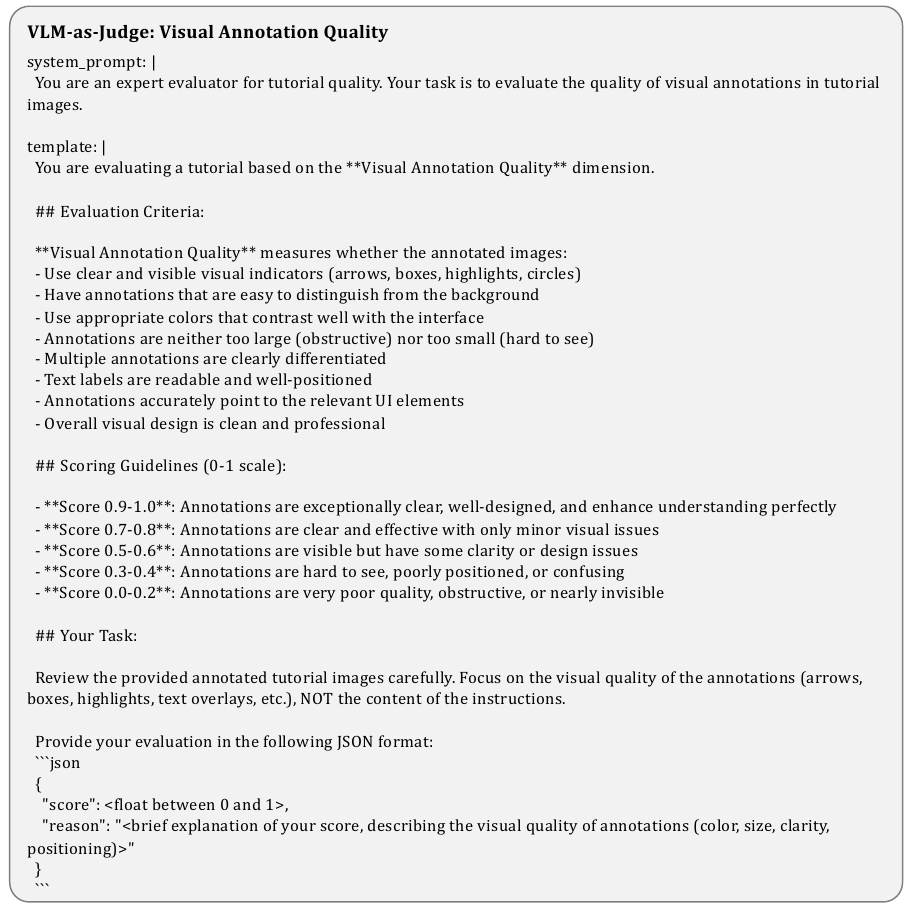}
    \caption{VLM-as-Judge prompt for Annotation Quality.}
    \label{fig:vlm_judge_prompt_anno_quality}
\end{figure*}

\begin{figure*}[t!]
    \centering
    \includegraphics[width=0.97\textwidth]{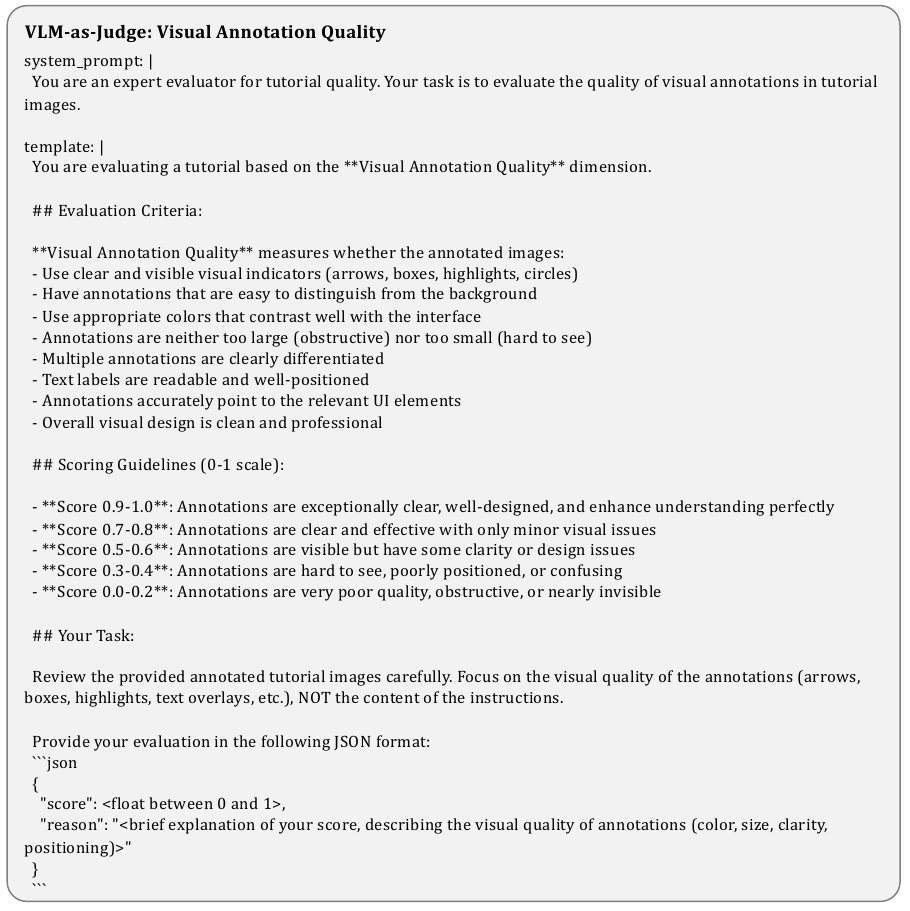}
    \caption{VLM-as-Judge prompt for Image Relevance.}
    \label{fig:vlm_judge_prompt_img_relevance}
\end{figure*}

\textit{Visual Score} evaluates the effectiveness of visual components in supporting comprehension:
\begin{itemize}
    \item \textbf{Annotation Quality}: Are visual markers (arrows, circles, highlights, magnifiers) appropriately applied to guide user attention to relevant UI elements? See \cref{fig:vlm_judge_prompt_anno_quality} for the prompt.
    \item \textbf{Image Relevance}: Do the selected screenshots accurately correspond to the described actions and capture the relevant UI state? See \cref{fig:vlm_judge_prompt_img_relevance} for the prompt.
\end{itemize}

For each dimension, we prompt GPT-4o to provide a score from 0 to 1, along with brief justification.
The final Content Score and Visual Score are computed as the average of their respective dimensions, and the Overall Score is the average across the five dimensions.

\paragraph{Human Consistency Validation}
\label{sec:supp_vlm_validation}
To assess the validity of VLM-as-Judge as a scalable proxy for human judgment, we conduct a human consistency evaluation on a diverse subset of TutorialBench.
We first sample 3 tasks from each of the 7 software applications (\(7 \times 3 = 21\) tasks), and for each task collect three tutorial variants (Demo2Tutorial, official human tutorial, and a vision-based baseline), yielding \(21 \times 3 = 63\) tutorials in total.
Two independent annotators then score each tutorial on the same five dimensions as our VLM-as-Judge protocol following the guideline in Fig.~\ref{fig:human_eval_guide}.
For analysis, we compute each rater's overall score by averaging the five dimension scores, and report Spearman's rank correlation (\(\rho\)) to measure agreement between human raters and VLM-as-Judge.
As shown in Tab.~\ref{tab:supp_vlm_validation}, the two raters have strong agreement (\(\rho=0.601\)), and the averaged human scores correlate well with VLM-as-Judge (\(\rho=0.755\)), supporting the reliability of our evaluation protocol.

\begin{table}[t]
    \centering
    \small
    \setlength{\tabcolsep}{6pt}
    \caption{\textbf{Human consistency validation for VLM-as-Judge.} Two annotators independently score 63 tutorials, and we compute Spearman's $\rho$ between human and VLM-as-Judge scores.}
    \label{tab:supp_vlm_validation}
    \vspace{-2mm}
    \begin{tabular}{@{}lc@{}}
        \toprule
        \textbf{Metric} & \textbf{Spearman's $\rho$} \\
        \midrule
        Inter-Rater Agreement & 0.601 \\
        Rater-1 \textit{vs.} VLM-as-Judge & 0.675 \\
        Rater-2 \textit{vs.} VLM-as-Judge & 0.671 \\
        Averaged Raters \textit{vs.} VLM-as-Judge & 0.755 \\
        \bottomrule
    \end{tabular}
    \vspace{-2mm}
\end{table}

\subsection{Runtime and Cost}
\label{sec:supp_runtime_cost}
We report the average runtime and estimated API cost of generating one tutorial using Demo2Tutorial.
We measure wall-clock time for the full generation pipeline after a demonstration is recorded (Action Parser \(\rightarrow\) Step Planner \(\rightarrow\) Tutorial Composer).
We also aggregate the total number of input/output tokens consumed across all model calls in the pipeline, and estimate API cost by summing token usage under the provider pricing used at the time of our experiments.
Tab.~\ref{tab:supp_runtime_cost} reports the results.

\begin{table}[t]
    \centering
    \scriptsize
    \setlength{\tabcolsep}{5pt}
    \caption{\textbf{Runtime and cost.} Average end-to-end runtime and estimated API cost per tutorial, averaged across TutorialBench.}
    \begin{tabular}{@{}cccc@{}}
        \toprule
        \textbf{Input Tok. (K)} & \textbf{Output Tok. (K)} & \textbf{Time (s)} & \textbf{Cost (USD)} \\
        \midrule
        259 & 3 & 368 & 1.36 \\
        \bottomrule
    \end{tabular}
    \vspace{-2mm}
    \label{tab:supp_runtime_cost}
    \vspace{-4mm}
\end{table}

\subsection{OSWorld Experiment}

To investigate whether multimodal tutorials can serve as effective external knowledge for enhancing GUI agent planning capabilities, we conduct experiments on the OSWorld benchmark~\cite{xie2024osworld}, focusing on two representative application domains: Chrome (web browser) and VLC (media player).

\begin{figure*}[t!]
    \centering
    \includegraphics[width=0.97\textwidth]{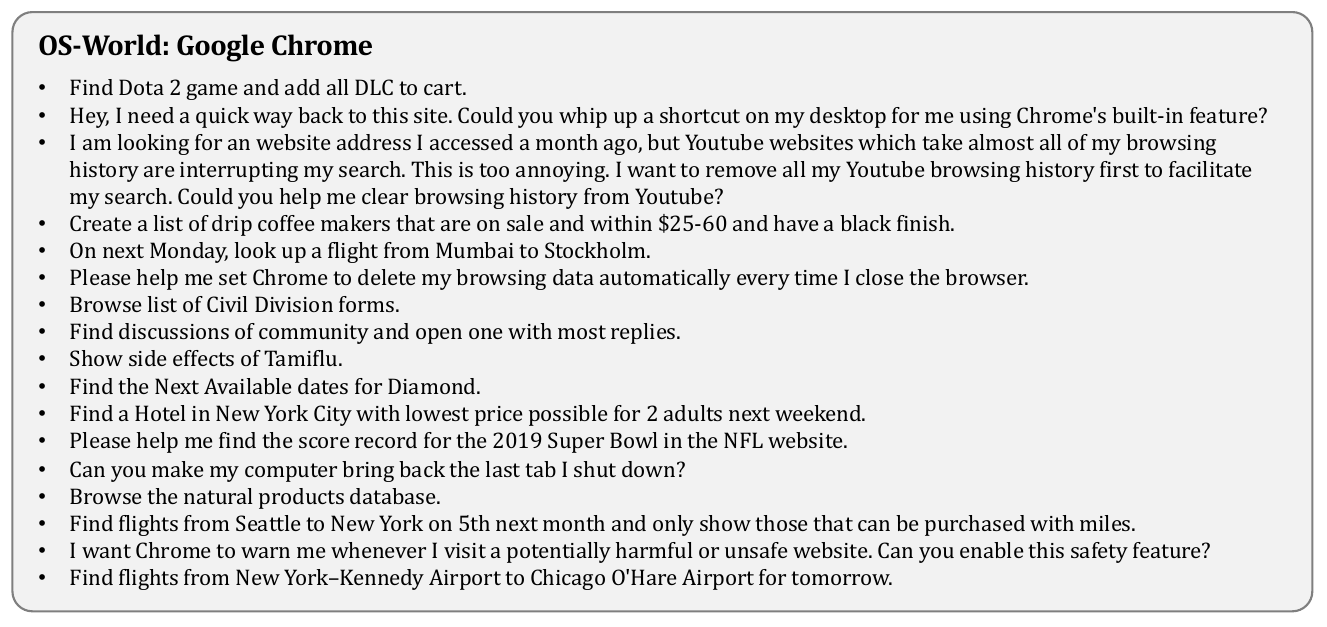}
    \caption{Task list of OS-World Chrome domain.}
    \label{fig:osworld_chrome}
\end{figure*}

\begin{figure*}[t!]
    \centering
    \includegraphics[width=0.97\textwidth]{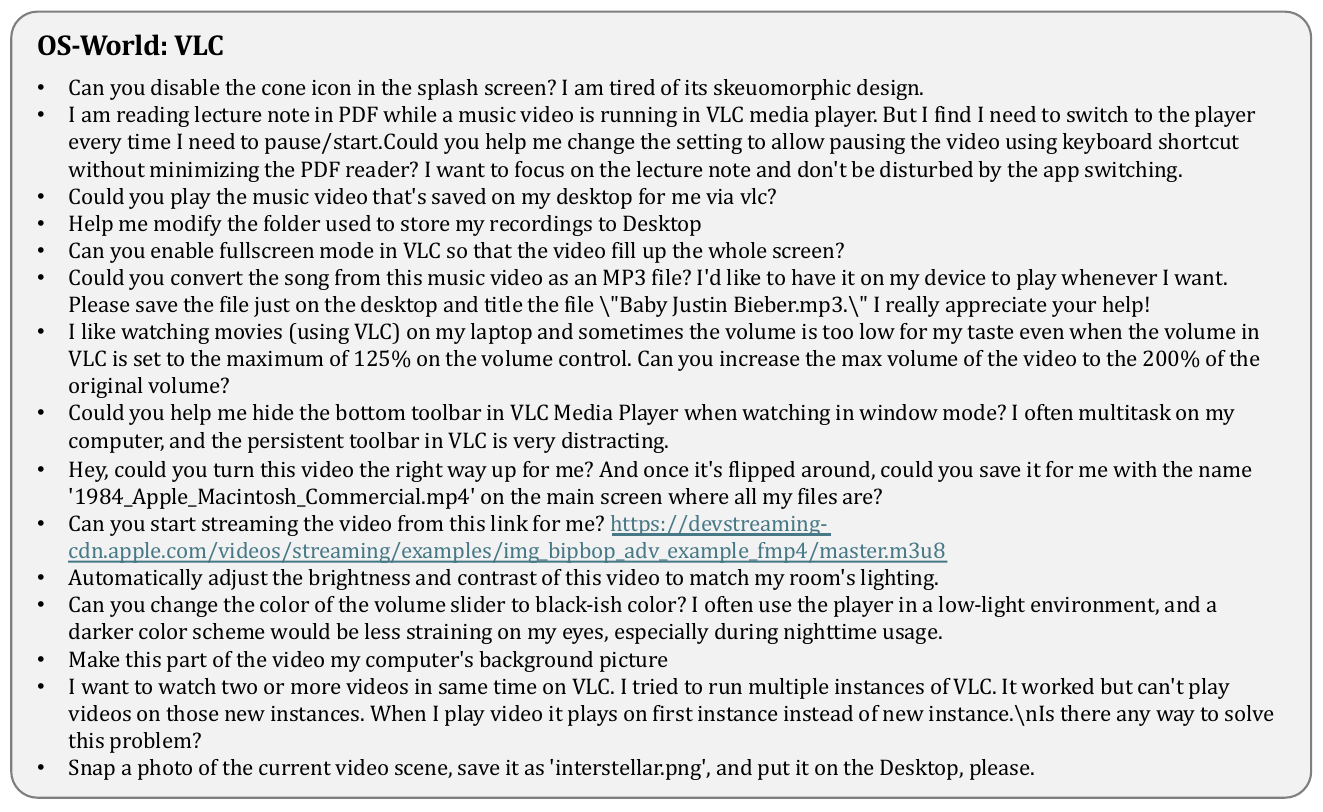}
    \caption{Task list of OS-World VLC domain.}
    \label{fig:osworld_vlc}
\end{figure*}

We recruit human experts familiar with Chrome and VLC to execute a curated set of tasks from the OSWorld benchmark suite.
Specifically, we select 17 tasks for Chrome and 14 tasks for VLC, covering diverse operations such as browser configuration, media playback settings, and interface customization.
The complete task lists are reported in \cref{fig:osworld_chrome} and \cref{fig:osworld_vlc}.
During task execution, the expert's interactions are captured using our HE-Recorder, which synchronously records screen video at 30 FPS and logs all low-level user actions (mouse clicks, keyboard inputs, window operations).
The recorded demonstrations are then processed through the full Demo2Tutorial pipeline to generate multimodal tutorials for each task.

We follow the Agent-S3~\cite{agents3} as the downstream GUI agent framework, which provides a modular architecture for planning, grounding and execution.
The generated tutorials are integrated as contextual knowledge into the agent's planner module.
We evaluate two strong planning models: \textit{o4-mini} and \textit{GPT-5}, both serving as the planning backbone within GUI Agent.
Performance is measured as the task success rate.
To analyze the contribution of different tutorial modalities, we conduct ablation studies comparing four levels of contextual supervision:
\begin{itemize}
    \item \textbf{Baseline}: Prompt-only, without any tutorial guidance.
    \item \textbf{+Text}: Incorporating textual step descriptions from tutorials.
    \item \textbf{+Image}: Incorporating visual screenshots from tutorials.
    \item \textbf{+Tutorial}: Incorporating full multimodal tutorials.
\end{itemize}
Each configuration is evaluated across all tasks within each domain, and results are averaged to obtain domain-specific success rates.

\begin{figure}[t!]
    \centering
    \includegraphics[width=\linewidth]{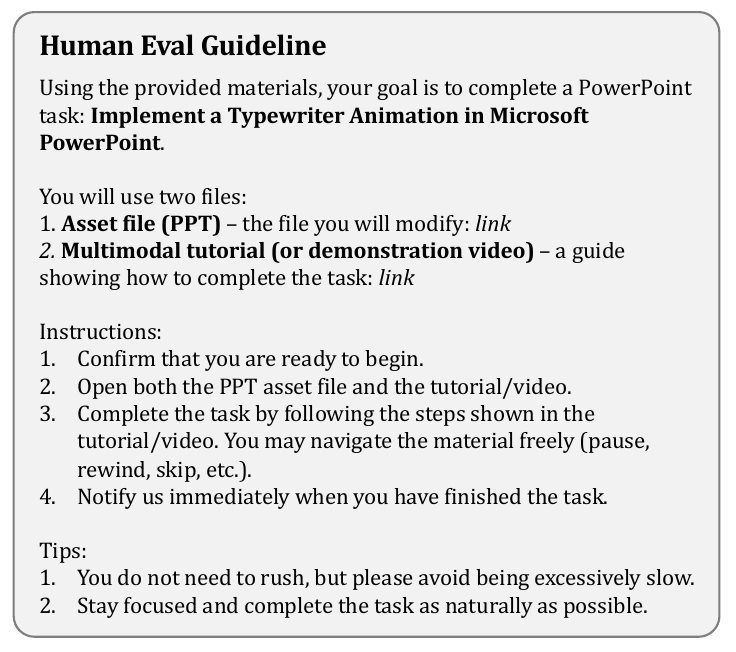}
    \caption{Human evaluation guideline.}
    \label{fig:human_eval_guide}
    \vspace{-10pt}
\end{figure}

\subsection{Human Evaluation}

We recruit 20 participants and randomly assign them into two groups (10 per condition).
The task involves implementing a seldom-used animation effect in Microsoft PowerPoint, specifically, creating a custom motion path animation with timing adjustments.
This task is intentionally chosen to be non-trivial, requiring multiple steps that are not immediately obvious to novice users.
Participants in the \textit{Demo Video} condition watch the raw screen recording demonstration, while those in the \textit{Tutorial} condition study the image-text interleaved tutorial generated by Demo2Tutorial.
After the learning phase, participants attempt to complete the task independently while being timed.
All participants successfully completed the task, confirming that both learning materials are effective.
The study protocol is shown in \cref{fig:human_eval_guide}.

\section{Qualitative Examples}

\subsection{Examples for Each Software}
\begin{figure*}[t!]
    \centering
    \includegraphics[width=0.95\textwidth]{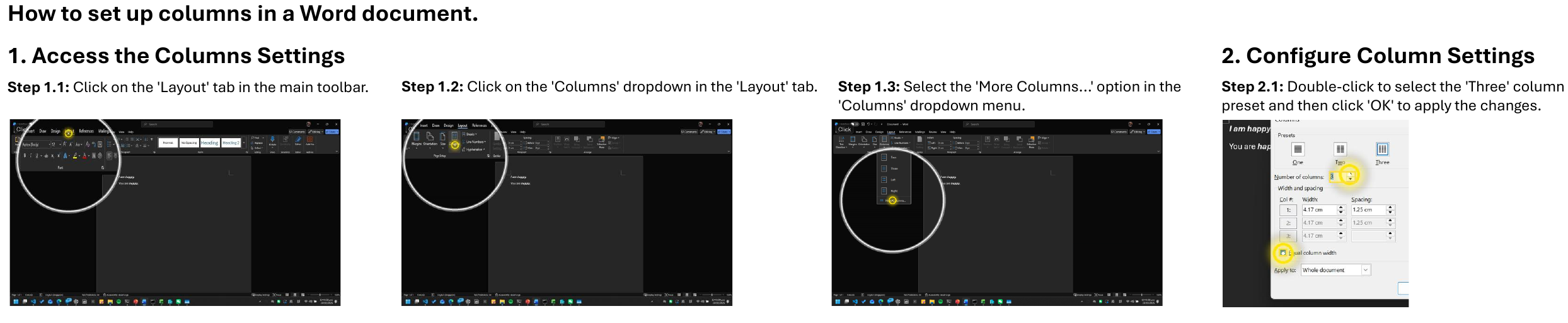}
    \caption{Example of Word tutorial.}
    \label{fig:word_tutorial}
\end{figure*}

\begin{figure*}[t!]
    \centering
    \includegraphics[width=0.75\textwidth]{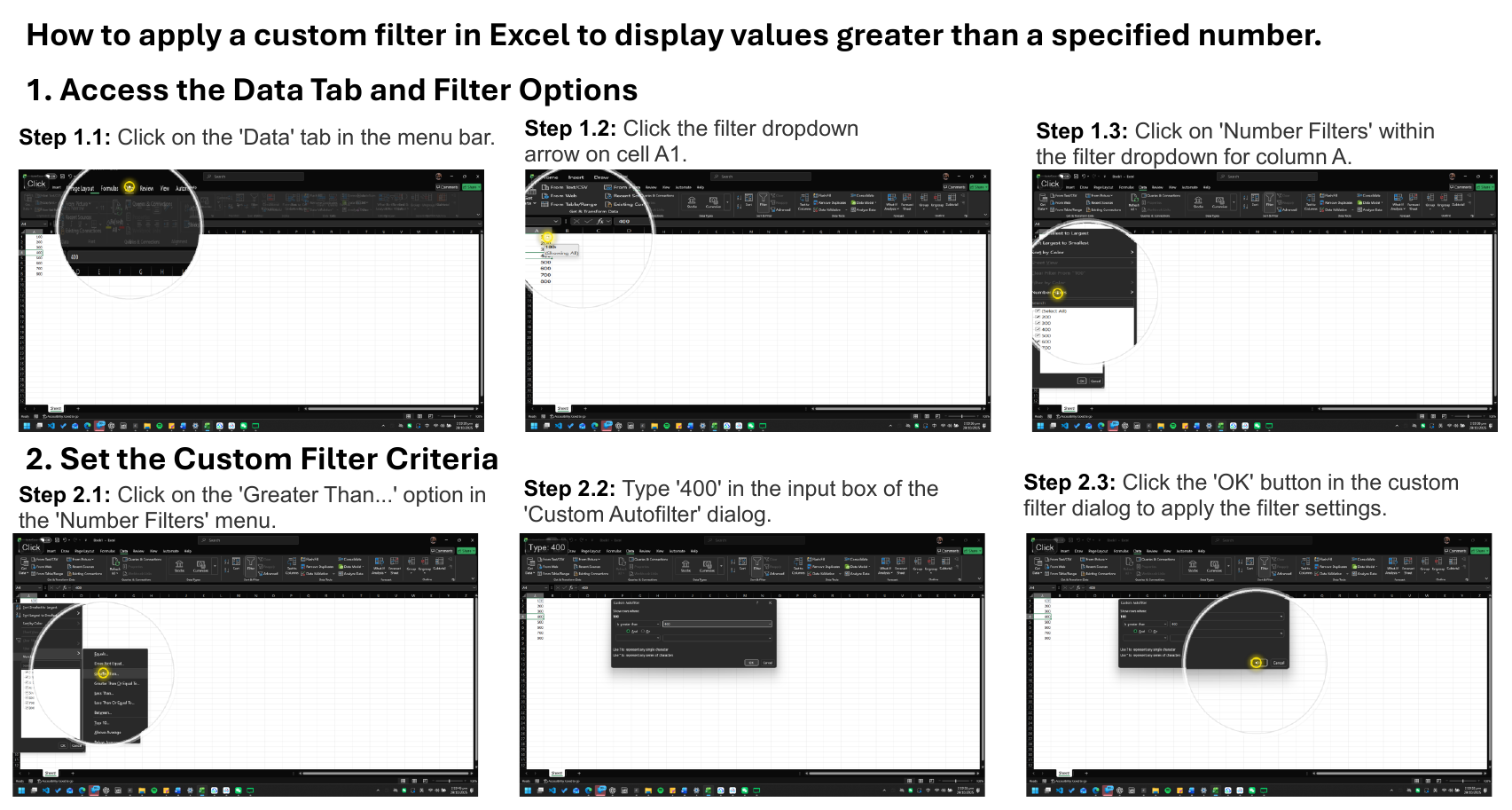}
    \caption{Example of Excel tutorial.}
    \label{fig:excel_tutorial}
\end{figure*}

\begin{figure*}[t!]
    \centering
    \includegraphics[width=0.75\textwidth]{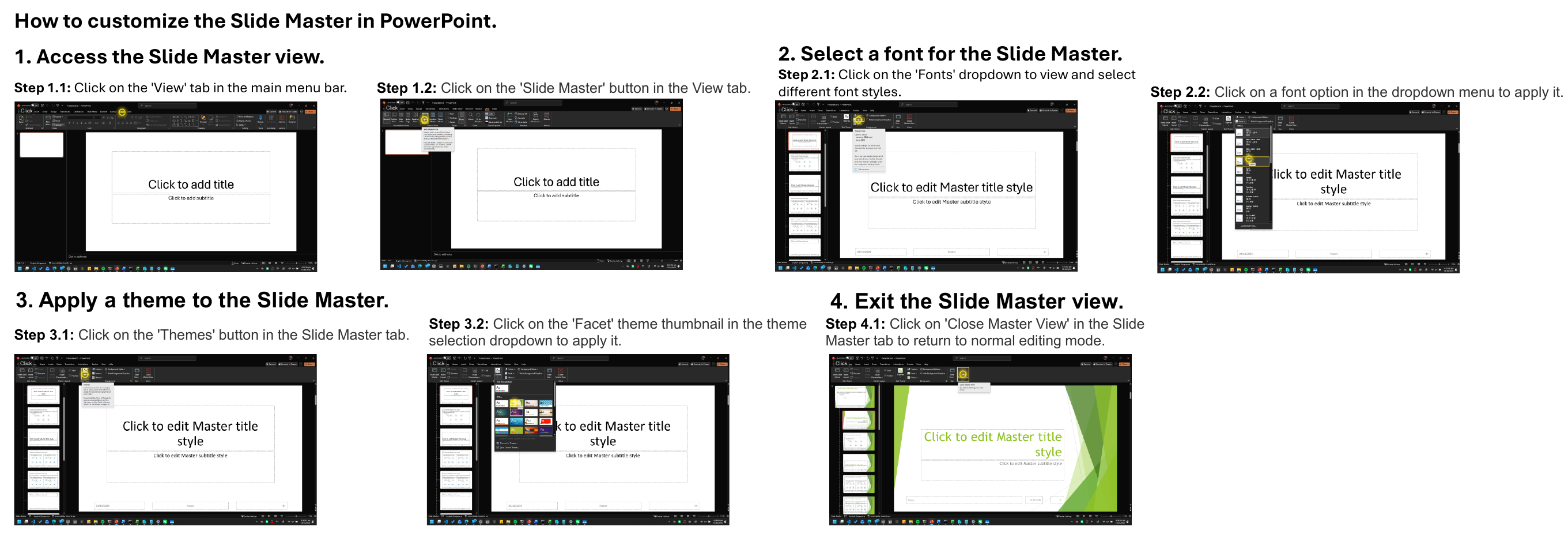}
    \caption{Example of PowerPoint tutorial.}
    \label{fig:ppt_tutorial}
\end{figure*}

\begin{figure*}[t!]
    \centering
    \includegraphics[width=0.75\textwidth]{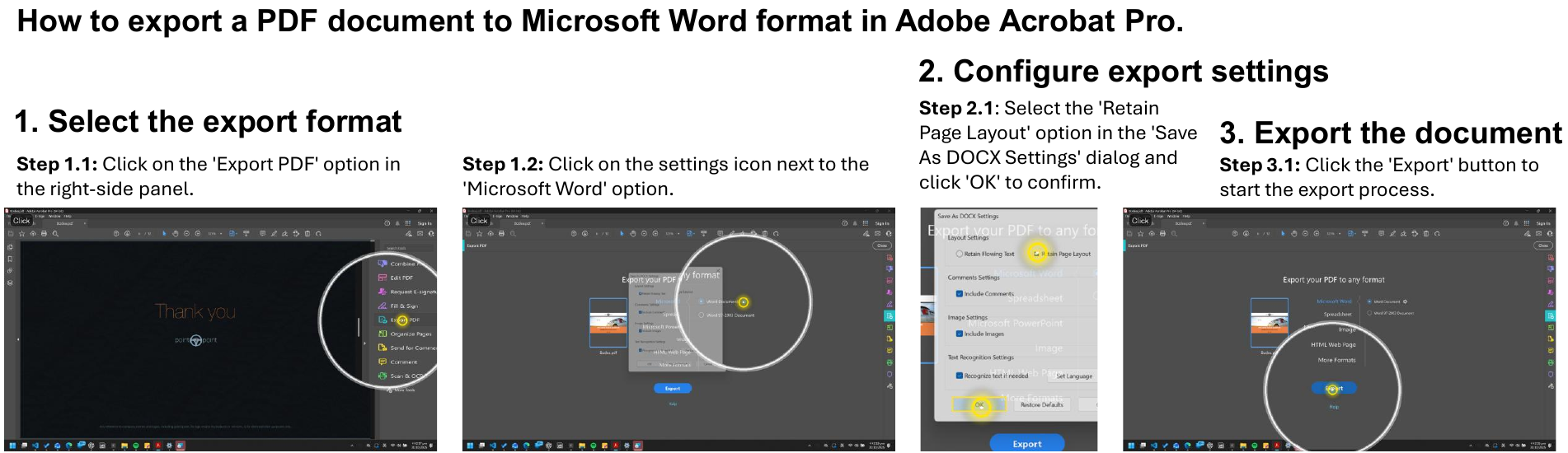}
    \caption{Example of Acrobat tutorial.}
    \label{fig:acrobat_tutorial}
\end{figure*}

\begin{figure*}[t!]
    \centering
    \includegraphics[width=0.95\textwidth]{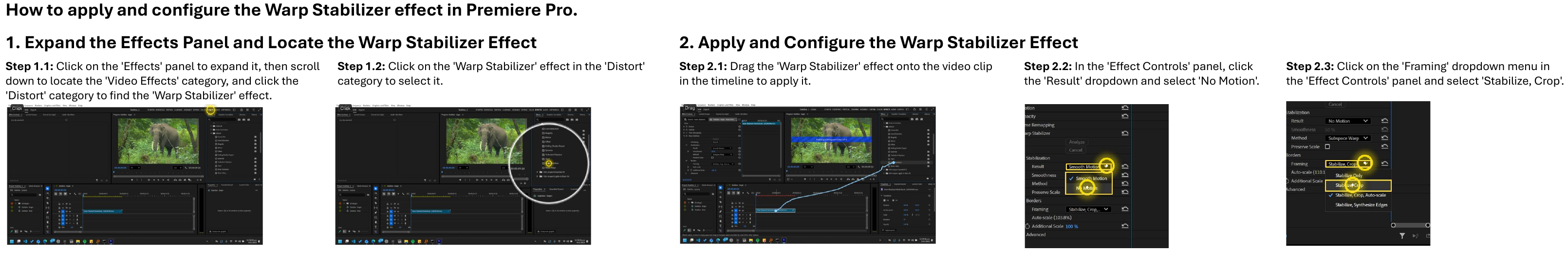}
    \caption{Example of Premiere Pro tutorial.}
    \label{fig:premiere_pro_tutorial}
\end{figure*}

\begin{figure*}[t!]
    \centering
    \includegraphics[width=0.95\textwidth]{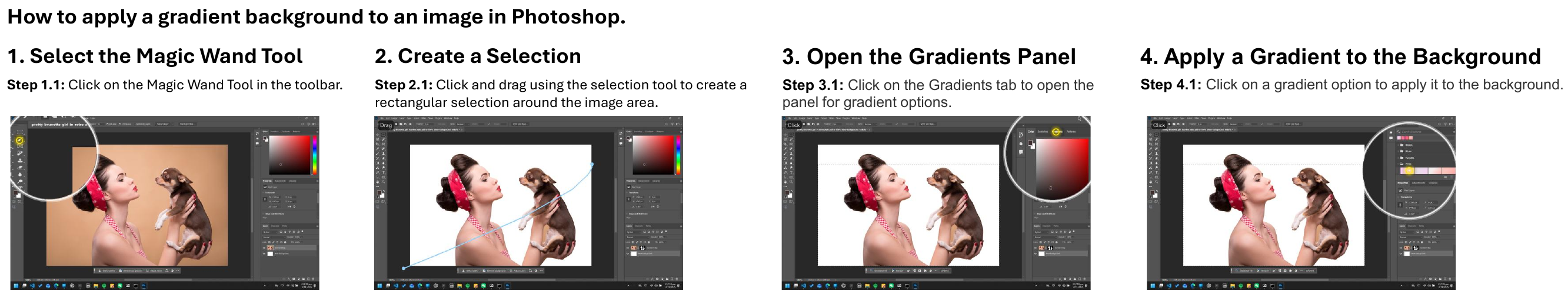}
    \caption{Example of Photoshop tutorial.}
    \label{fig:photoshop_tutorial}
\end{figure*}

In \cref{fig:word_tutorial,fig:excel_tutorial,fig:ppt_tutorial,fig:premiere_pro_tutorial,fig:photoshop_tutorial,fig:acrobat_tutorial}, we show qualitative examples of our generated tutorials across different software.

\subsection{Failure Cases}
\begin{figure*}[t!]
    \centering
    \includegraphics[width=0.85\textwidth]{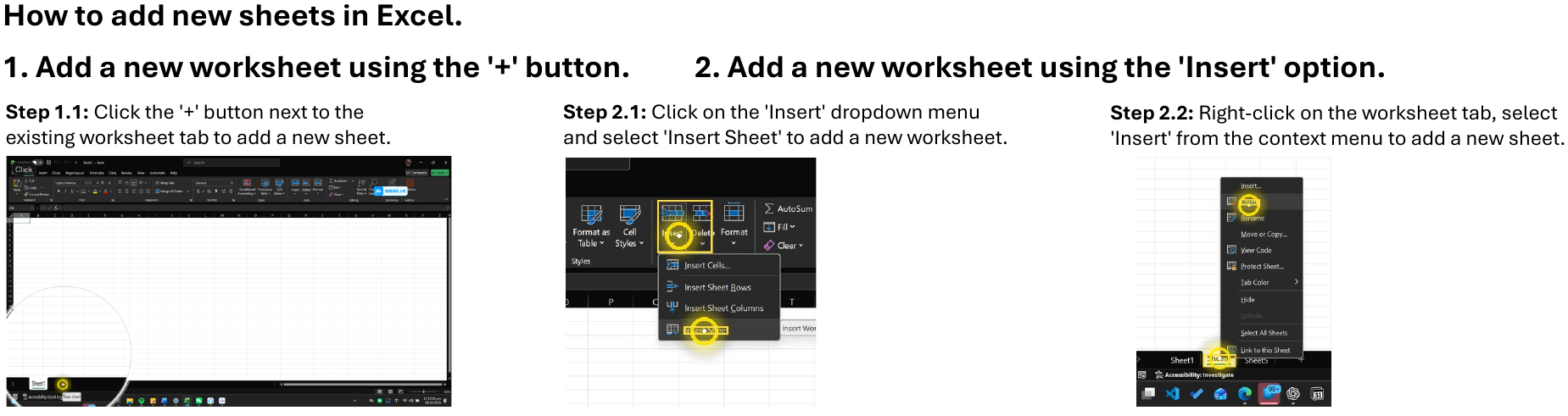}
    \caption{Example of failed Excel tutorial. The original video demonstration including both how to insert and delete a worksheet in Excel, but the generated tutorial only contains the instruction of inserting a worksheet.}
    \label{fig:failed_excel_tutorial}
\end{figure*}

\begin{figure*}[t!]
    \centering
    \includegraphics[width=0.75\textwidth]{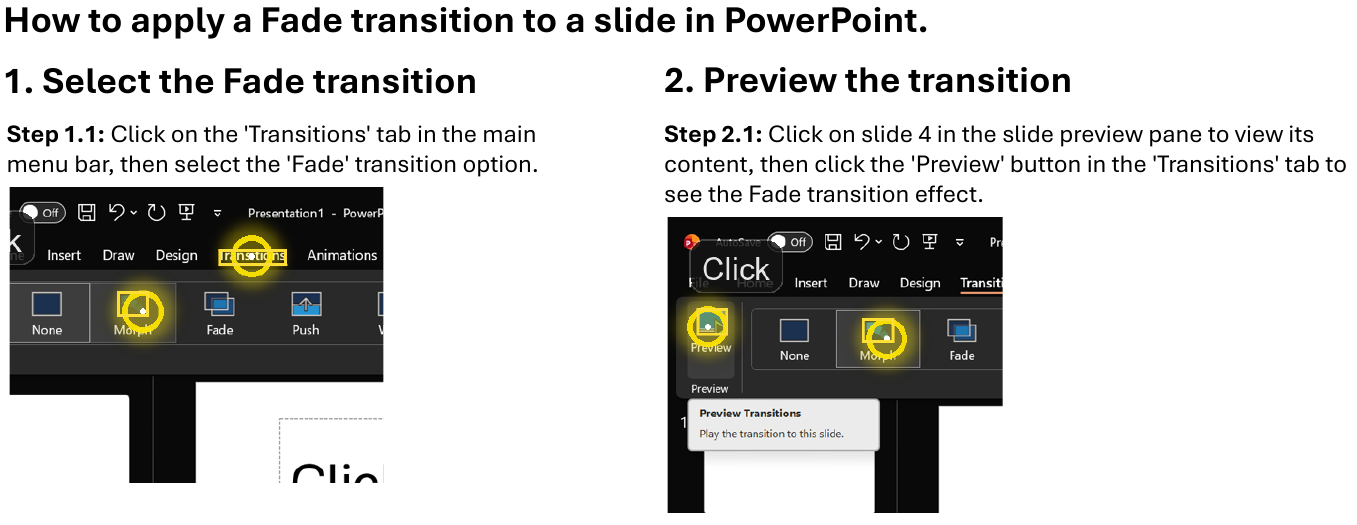}
    \caption{Example of failed PPT tutorial. The agent underlying MLLM fails to correctly recognize the action area and misinterprets the Morph animation as a Fade animation.}
    \label{fig:failed_ppt_tutorial}
\end{figure*}

Fig.~\ref{fig:failed_excel_tutorial} and Fig.~\ref{fig:failed_ppt_tutorial} show two representative failure cases of our generated tutorials.
First, Fig.~\ref{fig:failed_excel_tutorial} shows that the generated tutorial only contains the instruction text of inserting a worksheet, while the original video demonstration includes both how to insert and delete a worksheet, as reflected in the image.
This failure happens in the Planner stage, where the agent tries to condense the raw actions into hierarchical task graphs representing goals and steps, but overlooks the deletion operations.
Second, Fig.~\ref{fig:failed_ppt_tutorial} shows that the visual guidance is inconsistent with the text description.
The screenshots show the Morph animation, but the text description is about the Fade animation.
This failure happens in the Action Parser stage, where the agent tries to parse the raw actions into semantic descriptions, but fails to correctly recognize the action area and misinterprets the Morph animation as a Fade animation.

\section{Future Works}

Building upon the current Demo2Tutorial framework, we identify several promising research directions that could further enhance the framework's capabilities and broaden its applicability.

While the current framework leverages HE-Recorder's synchronized action logs for precise temporal grounding, an exciting future direction is to develop vision-based action inference models that can automatically reconstruct low-level operations (clicks, keystrokes, gestures) purely from consecutive video frames.
Such capability would enable the framework to process arbitrary screen recordings from the internet, dramatically expanding the scalability and applicability of tutorial generation to the vast repository of existing demonstration videos without requiring specialized recording tools.
This research direction would bridge computer vision and human-computer interaction, advancing the state-of-the-art in visual action understanding.

The Demo2Tutorial pipeline achieves substantial knowledge compression (325.71× on average), transforming lengthy demonstrations into concise tutorials.
A promising future direction is to develop intelligent pre-screening mechanisms that can identify high-information segments before the full parsing stage, enabling adaptive processing strategies.
For instance, visual similarity detection could cluster redundant frames, while saliency estimation could prioritize segments with significant UI state changes.
Such mechanisms would reduce computational overhead for long workflows while preserving tutorial quality, making the framework more efficient and cost-effective for processing extensive demonstration videos.

\end{document}